\documentclass[final,1p,times]{elsarticle}

\usepackage[T5]{fontenc}

\usepackage{times}
\usepackage{latexsym}

\usepackage{array}
\usepackage{graphicx}
\usepackage{multirow}
\usepackage{pgf-pie}   
\usepackage{soul}
\usepackage{color}
\usepackage{tablefootnote}
\usepackage{adjustbox}
\usepackage{url}
\usepackage{float}
\usepackage{graphicx,graphics}
\usepackage{caption}
\usepackage{subcaption}
\usepackage[bookmarks,bookmarksopen,bookmarksdepth=3, bookmarksnumbered]{hyperref}
\usepackage{longtable}
\usepackage{amsmath}
\usepackage{amssymb}
\usepackage{pifont}
\usepackage{xcolor}
\usepackage{algorithm}
\usepackage{algpseudocode}

\usepackage{lipsum}

\usepackage{xcolor}
\usepackage{soul, soulutf8}

\usepackage{scalerel,graphicx,xparse}

\NewDocumentCommand\emojione{}{\scalerel*{\includegraphics{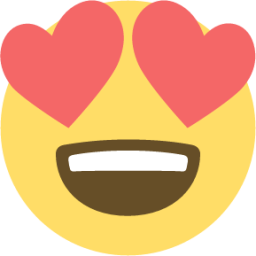}}{X}}
\NewDocumentCommand\emojitwo{}{\scalerel*{\includegraphics{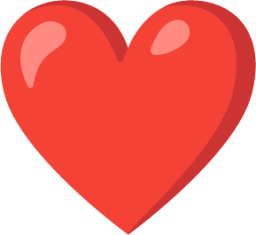}}{X}}
\NewDocumentCommand\emojithree{}{\scalerel*{\includegraphics{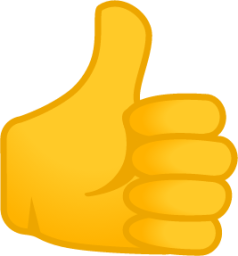}}{X}}

\definecolor{lavender}{RGB}{240,230,250}
\definecolor{lightblue}{RGB}{173, 216, 230}

\sethlcolor{lightgray}

\newcommand{\cmark}{\ding{51}}%
\newcommand{\xmark}{\ding{55}}%

\begin{document}
\let\WriteBookmarks\relax
\def\floatpagepagefraction{1}
\def\textpagefraction{.001}

% Main title  of the paper
\title{New Benchmark Dataset and Fine-Grained Cross-Modal Fusion Framework for Vietnamese Multimodal Aspect-Category Sentiment Analysis}

%fine-grained cross-modal fusion
% First author/
\author[1,2]{Quy Hoang Nguyen}
\ead{20521815@gm.uit.edu.vn}

\author[1,2]{Minh-Van Truong Nguyen}
\ead{20522146@gm.uit.edu.vn}

\author[1,2]{Kiet Van Nguyen\corref{cor1}}%[orcid=0000-0002-8456-2742]
\ead{kietnv@uit.edu.vn}

\affiliation[1]{organization={Faculty of Information Science and Engineering, University of Information Technology\\},
    city={Ho Chi Minh City},
    %postcode={70000}, 
    country={Vietnam}}
    
\affiliation[2]{organization={Vietnam National University},
    city={Ho Chi Minh City},
    %postcode={70000}, 
    country={Vietnam}}

\cortext[cor1]{Corresponding author}

\begin{abstract}

The emergence of multimodal data on social media platforms presents new opportunities to better understand user sentiments toward a given aspect. However, existing multimodal datasets for Aspect-Category Sentiment Analysis (ACSA) often focus on textual annotations, neglecting fine-grained information in images. Consequently, these datasets fail to fully exploit the richness inherent in multimodal. To address this, we introduce a new Vietnamese multimodal dataset, named ViMACSA, which consists of 4,876 text-image pairs with 14,618 fine-grained annotations for both text and image in the hotel domain. Additionally, we propose a Fine-Grained Cross-Modal Fusion Framework (FCMF) that effectively learns both intra- and inter-modality interactions and then fuses these information to produce a unified multimodal representation. Experimental results show that our framework outperforms SOTA models on the ViMACSA dataset, achieving the highest F1 score of 79.73\%. We also explore characteristics and challenges in Vietnamese multimodal sentiment analysis, including misspellings, abbreviations, and the complexities of the Vietnamese language. This work contributes both a benchmark dataset and a new framework that leverages fine-grained multimodal information to improve multimodal aspect-category sentiment analysis. Our dataset is available for research purposes.\footnote{\href{https://github.com/hoangquy18/Multimodal-Aspect-Category-Sentiment-Analysis.git}{https://github.com/ViMACSA}}

\end{abstract}

%    \item Creating a new Vietnamese dataset for {\bf multimodal} aspect-category sentiment analysis in the hotel domain.
%    \item Proposing {\bf Fine-Grained Cross-Modal Fusion Framework (FCMF)} that learns the interactions between fine-grained elements derived from text and images.
%    \item Providing a {\bf comprehensive experiment} on multimodal aspect-category sentiment analysis in Vietnamese.
%    \item Investigating {\bf challenges and characteristics} in Vietnamese multimodal sentiment analysis tasks.
    

% Keywords

\maketitle

\section{Introduction} \label{Introduction}
% todo: fine-grained 
In the ever-changing service industry, especially in hospitality, focusing on user experience is vital for businesses. An effective way to achieve this is to analyze the comments and feedback provided by users and make the necessary adjustments accordingly. Sentiment analysis is a popular topic in natural language processing that has attracted significant attention from researchers due to its diverse development.

Aspect-Based Sentiment Analysis (ABSA) is an advanced task within sentiment analysis. Its purpose is to classify sentiments based on the aspects identified within a sentence. Given a sentence as input, the output includes identified aspects and their corresponding sentiment. ABSA plays a crucial role in bridging the communication gap between customers and businesses by helping companies better understand their customers' needs and preferences.

The emergence of multimodal approaches has led to new research directions for various tasks, including ABSA. Multimodal Aspect-Based Sentiment Analysis (MABSA) aims to integrate relevant data beyond text to determine aspect-oriented sentiments within a given text \cite{das2023multimodal,zhang2022survey}. This approach distinguishes from traditional sentiment analysis by incorporating additional modalities to capture user-related information that may not be explicitly mentioned in the text. 

\begin{figure}[H]
    \centering
    \includegraphics[,height=7cm]{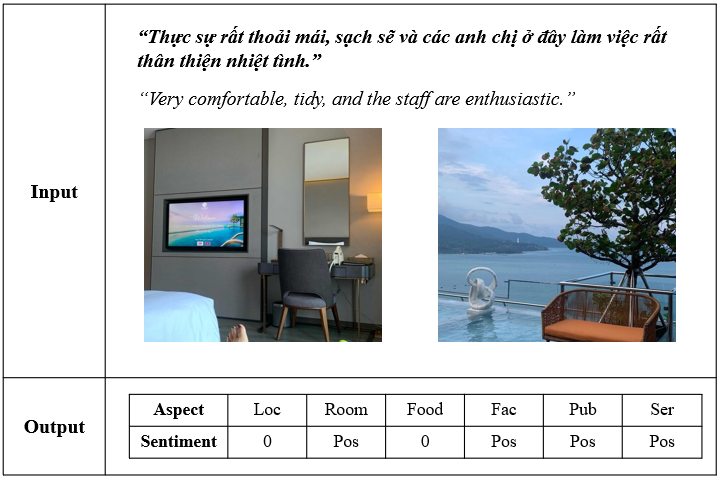}
    \caption{Examples of the Multimodal ACSA task in Vietnamese.}
    \label{fig:example_data}
\end{figure}

Social media platforms and review websites now allow users to upload images along with their comments. These images can be a valuable data source for multimodal ABSA. In Figure \ref{fig:example_data}, consider a user review saying, "Very comfortable, tidy, and the staff are enthusiastic". Here, the phrase "the staff are very enthusiastic" clearly conveys a positive sentiment towards the "staff" aspect. However, the phrase "very comfortable, tidy" expresses a positive sentiment but does not specify the aspect. This common scenario in online reviews makes it difficult for traditional ABSA approaches to identify the aspect precisely. To address this challenge, we introduce the task of Multimodal Aspect-Category Sentiment Analysis (MACSA), which leverages both text and images to determine the sentiment of an aspect category. For example, if the user provides photos of the hotel room, we can infer that "very comfortable, tidy" refers to the room itself. Such additional information can significantly enhance the performance of ABSA tasks.

The fine-grained elements in both text and images are essential as they provide crucial and interpretable cues. While text provides significant information, images also offer complementary information that enhances understanding. Therefore, annotating image data with fine-grained annotations is essential. In addition to creating a new dataset, we propose a framework called Fine-Grained Cross-Modal Fusion (FCMF), which learns both intra- and inter-modality interactions between textual and visual elements. Our article aims to make the following contributions. 

%Viết lại cái highlights nhưng chi tiết hơn chút.
\begin{itemize}
    \item We introduce a Vietnamese multimodal dataset with fine-grained annotations for the MACSA task, named ViMACSA. This new benchmark dataset contains fine-grained image annotations.
        
    \item We propose a novel framework called FCMF based on Attention Mechanisms. This framework learns the interactions between fine-grained elements derived from text and images.

    \item We evaluate our framework against several baseline models on the ViMACSA dataset. Experimental results show that our framework outperforms baseline models and demonstrates its potential to support baseline comparisons in future research using this dataset.

    \item We investigate challenges and characteristics in Vietnamese multimodal sentiment analysis tasks, with an emphasis on common Vietnamese language errors such as misspellings, abbreviations, and the complexities of Vietnamese language processing.

\end{itemize}

% TODO
%Thấy Kiệt: Thêm 01 đoạn giới thiệu về bố cục bài báo. For example: The remainder of this paper is organized as follows. Section 2 reviews the literature of multimodal sentiment analysis. Section 3 describes the extension of the multimodal sentiment datasets. Section 4 describes the proposed model in detail. Section 5 reports the settings and results of the experiments. Section 6 concludes the paper.

The remainder of this article is organized as follows. Section \ref{sec:related_work} provides a summary of related work. Section \ref{sec:masc_dataset} presents the definition of the MACSA task, the process of creating the dataset, the annotation guidelines, and the dataset evaluation. Section \ref{sec:proposed_fcmf} describes the proposed framework in detail. The experimental results are analyzed in Section \ref{sec:experiment}. Finally, Section \ref{sec:conclusions} concludes the article and provides directions for future work.

\section{Related Work} \label{sec:related_work}

Multimodal has received extensive attention from researchers worldwide \cite{lin2024adapt,bayoudh2023survey,li2024multimodal,fan2024transformer,geetha2024multimodal,qu2024qmfnd}, including the field of multimodal sentiment analysis. Multimodal sentiment analysis (MSA) aims to understand emotions conveyed through a combination of modalities, such as speech, images, and text. Researchers have developed various datasets containing multimodal information to support MSA research (Section \ref{sec:msa_dataset}). Different methods have emerged to solve MSA, such as LSTM and BERT-based models, which allow a nuanced analysis of complex interactions between these modalities (Section \ref{sec:msa_method}). Furthermore, we comprehensively review the existing datasets and methods available for processing multimodal data in Vietnamese (Section \ref{sec:msa_vn}). 

\subsection{Multimodal Sentiment Analysis Dataset}
\label{sec:msa_dataset}

With the advancement of technology, people can express their feelings through audio and visual modalities. Multimodal sentiment analysis takes advantage of these new possibilities, offering a powerful extension of traditional text-based analysis by incorporating insights from visual and audio data. This field has rapidly gained momentum, attracting researchers worldwide \cite{zhao2024survey}. Numerous high-quality datasets are now available to support this research, as shown in Table \ref{tab:refer}. Most of them focus on video-based sentiment datasets, including \cite{busso2008iemocap, koelstra2011deap, zadeh2016mosi,zadeh2018multimodal,poria2018meld,yu2020ch,zadeh2020cmu}

However, there are relatively few multimodal sentiment analysis datasets that focus on text-image data. In 2019, Cai et al. \cite{cai2019multi} introduced a multimodal dataset for sarcasm classification. In 2021, Zhou et al. \cite{zhou2021masad} created a multimodal dataset with 38,532 samples that covered seven domains and 57 aspects. In 2022, Ramamoorthy et al. \cite{ramamoorthy2022memotion} developed a multimodal meme-based dataset with 10,000 samples for three tasks: Sentiment Analysis, Emotion Classification, and Intensity of Emotion Classes. 

All of these datasets are annotated for sentiment polarity at the sentence level. For datasets focusing on aspect-level tasks, Xu et al. \cite{xu2019multi} released a multimodal dataset containing 5,528 samples in the mobile phone domain, annotated on six aspects (i.e. screen, photographing effect, appearance and feeling, performance configuration, battery life, and price-performance ratio). In 2019, Yu et al. \cite{yu2019adapting} released two datasets, Twitter-15 and Twitter-17, which are annotated sentiment polarity towards each target.

\begin{table}[h]
\centering
\caption{Several benchmark datasets exist for Multimodal Sentiment Analysis. A, V, and T denote audio, visual, and textual modalities, respectively.}
\label{tab:refer}
\resizebox{\columnwidth}{!}{
\begin{tabular}{lllllccclll}
\multirow{3}{*}{Dataset} & \multirow{3}{*}{Year} & \multirow{3}{*}{Source} & \multirow{3}{*}{Modality} & \multirow{3}{*}{Language} & \multicolumn{3}{c}{Sentiments} & \multirow{3}{*}{Labels} & \multirow{3}{*}{\#Aspects} & \multirow{3}{*}{\#Samples} \\ \cline{6-8}
 &  &  &  &  & \multirow{2}{*}{T} & \multirow{2}{*}{V} & \multirow{2}{*}{Multimodal} &  &  &  \\
 &  &  &  &  &  &  &  &  &  &  \\ \hline
 
\multirow{3}{*}{MVSA-Multiple \cite{MVSA}} & \multirow{3}{*}{2016} & \multirow{3}{*}{Twitter} & \multirow{3}{*}{V+T} & \multirow{3}{*}{English} & \multicolumn{1}{l}{\multirow{3}{*}{\cmark}} & \multicolumn{1}{l}{\multirow{3}{*}{\cmark}} & \multicolumn{1}{l}{\multirow{3}{*}{}} & \multirow{3}{*}{Neg, Neu, Pos} & \multirow{3}{*}{-} & \multirow{3}{*}{19,600} \\
 &  &  &  &  & \multicolumn{1}{l}{} & \multicolumn{1}{l}{} & \multicolumn{1}{l}{} &  &  &  \\
 &  &  &  &  & \multicolumn{1}{l}{} & \multicolumn{1}{l}{} & \multicolumn{1}{l}{} &  &  &  \\ \hline
 
\multirow{3}{*}{B-T4SA \cite{vadicamo2017cross}} & \multirow{3}{*}{2017} & \multirow{3}{*}{Twitter} & \multirow{3}{*}{V+T} & \multirow{3}{*}{English} & \multirow{3}{*}{\cmark} & \multirow{3}{*}{} & \multirow{3}{*}{} & \multirow{3}{*}{[-3,3]} & \multirow{3}{*}{-} & \multirow{3}{*}{470,586} \\
 &  &  &  &  &  &  &  &  &  &  \\
 &  &  &  &  &  &  &  &  &  &  \\ \hline

\multirow{3}{*}{CMU-MOSEI \cite{zadeh2018multimodal}} & \multirow{3}{*}{2018} & \multirow{3}{*}{Youtube} & \multirow{3}{*}{ A+V+T} & \multirow{3}{*}{English} & \multirow{3}{*}{} & \multirow{3}{*}{} & \multirow{3}{*}{\cmark} & \multirow{3}{*}{[-3,3]} & \multirow{3}{*}{-} & \multirow{3}{*}{23,453} \\
&  &  &  &  &  &  &  &  &  &  \\
&  &  &  &  &  &  &  &  &  &  \\ \hline

\multirow{3}{*}{Twitter-Sarcasm \cite{cai2019multi}} & \multirow{3}{*}{2019} & \multirow{3}{*}{Twitter} & \multirow{3}{*}{V+T} & \multirow{3}{*}{English} & \multirow{3}{*}{} & \multirow{3}{*}{} & \multirow{3}{*}{\cmark} & \multirow{3}{*}{Neg, Pos} & \multirow{3}{*}{-} & \multirow{3}{*}{24,635} \\
&  &  &  &  &  &  &  &  &  &  \\
&  &  &  &  &  &  &  &  &  &  \\ \hline

\multirow{3}{*}{Twitter 15 \cite{yu2019adapting}} & \multirow{3}{*}{2019} & \multirow{3}{*}{Twitter} & \multirow{3}{*}{V+T} & \multirow{3}{*}{English} & \multirow{3}{*}{} & \multirow{3}{*}{} & \multirow{3}{*}{\cmark} & \multirow{3}{*}{Neg,Neu,Pos} & \multirow{3}{*}{-} & \multirow{3}{*}{5,338} \\
&  &  &  &  &  &  &  &  &  &  \\
&  &  &  &  &  &  &  &  &  &  \\ 

\multirow{3}{*}{Twitter 17 \cite{yu2019adapting}} & \multirow{3}{*}{2019} & \multirow{3}{*}{Twitter} & \multirow{3}{*}{V+T} & \multirow{3}{*}{English} & \multirow{3}{*}{} & \multirow{3}{*}{} & \multirow{3}{*}{\cmark} & \multirow{3}{*}{Neg,Neu,Pos} & \multirow{3}{*}{-} & \multirow{3}{*}{5,972} \\
&  &  &  &  &  &  &  &  &  &  \\
&  &  &  &  &  &  &  &  &  &  \\ \hline

\multirow{3}{*}{Multi-ZOL \cite{xu2019multi}} & \multirow{3}{*}{2019} & \multirow{3}{*}{ZOL} & \multirow{3}{*}{V+T} & \multirow{3}{*}{Chinese} & \multirow{3}{*}{} & \multirow{3}{*}{} & \multirow{3}{*}{\cmark} & \multirow{3}{*}{[1,10]} & \multirow{3}{*}{6} & \multirow{3}{*}{5,228} \\
&  &  &  &  &  &  &  &  &  &  \\
&  &  &  &  &  &  &  &  &  &  \\ \hline

\multirow{3}{*}{MELD \cite{poria2018meld}} & \multirow{3}{*}{2019} & \multirow{3}{*}{The Friends} & \multirow{3}{*}{A+V+T } & \multirow{3}{*}{English} & \multirow{3}{*}{} & \multirow{3}{*}{} & \multirow{3}{*}{\cmark} & \multirow{3}{*}{Emotions} & \multirow{3}{*}{7} & \multirow{3}{*}{13,000} \\
&  &  &  &  &  &  &  &  &  &  \\
&  &  &  &  &  &  &  &  &  &  \\ \hline

\multirow{3}{*}{CH-SIMS \cite{yu2020ch}} & \multirow{3}{*}{2020} & \multirow{3}{*}{Movie, Tv series, etc} & \multirow{3}{*}{A+V+T } & \multirow{3}{*}{Chinese} & \multirow{3}{*}{\cmark} & \multirow{3}{*}{\cmark} & \multirow{3}{*}{\cmark} & \multirow{3}{*}{[-1,1]} & \multirow{3}{*}{-} & \multirow{3}{*}{2,281} \\
&  &  &  &  &  &  &  &  &  &  \\
&  &  &  &  &  &  &  &  &  &  \\ \hline

\multirow{3}{*}{MASAD \cite{zhou2021masad}} & \multirow{3}{*}{2021} & \multirow{3}{*}{Yahoo!, Flickr} & \multirow{3}{*}{V+T } & \multirow{3}{*}{English} & \multirow{3}{*}{} & \multirow{3}{*}{} & \multirow{3}{*}{\cmark} & \multirow{3}{*}{Neg,  Pos} & \multirow{3}{*}{57} & \multirow{3}{*}{38,532} \\
&  &  &  &  &  &  &  &  &  &  \\
&  &  &  &  &  &  &  &  &  &  \\ \hline

\multirow{3}{*}{CMU-MOSEAS \cite{zadeh2020cmu}} & \multirow{3}{*}{2021} & \multirow{3}{*}{Youtube} & \multirow{3}{*}{A+V+T } & \multirow{3}{*}{Spanish, Portuguese,   German, French} & \multirow{3}{*}{} & \multirow{3}{*}{} & \multirow{3}{*}{\cmark} & \multirow{3}{*}{[-3,3], [0,3]} & \multirow{3}{*}{-} & \multirow{3}{*}{40,000} \\
&  &  &  &  &  &  &  &  &  &  \\
&  &  &  &  &  &  &  &  &  &  \\ \hline

\multirow{3}{*}{Memotion 2 \cite{ramamoorthy2022memotion}} & \multirow{3}{*}{2022} & \multirow{3}{*}{Facebook, Reddit, etc.} & \multirow{3}{*}{ V+T } & \multirow{3}{*}{English} & \multirow{3}{*}{\cmark} & \multirow{3}{*}{\cmark} & \multirow{3}{*}{\cmark} & \multirow{3}{*}{Neg, Neu, Pos, Emotions} & \multirow{3}{*}{20} & \multirow{3}{*}{10,000} \\
&  &  &  &  &  &  &  &  &  &  \\
&  &  &  &  &  &  &  &  &  &  \\ \hline

\end{tabular}}
\end{table}

\subsection{Multimodal Sentiment Analysis Method} \label{sec:msa_method}

Previous research on Aspect-Based Sentiment Analysis has mainly focused on sentiment analysis in text-based data, including \cite{hoang2019aspect,zhu2019aspect,liu2021solving,bu2021asap,liao2021improved}. However, for multimodal data, the objective is to identify aspect-sentiment relationships and fuse different modalities efficiently \cite{zhang2022survey,das2023multimodal,zhao2024survey,liu2024sarcasm,xiao2024atlantis,zhu2023skeafn,lu2024fact}. 

The existing fusion techniques can be classified into three main types: early fusion, intermediate fusion, and late fusion \cite{xu2019sentiment}. In early fusion, the features of various modalities concatenate into a joint feature representation. On the other hand, late fusion creates independent models for each modality, which are then integrated to produce the final output. However, these fusions cannot fully capture the interactions between modalities. 

To address this issue, various models using intermediate fusion have emerged to allow a more nuanced analysis of complex interactions across these modalities. For instance, in 2019, Xu et al. \cite{xu2019multi} introduced the MIMN model using interactive memory networks to learn cross-modality and self-modality interactions. In the same year, Yu et al. \cite{yu2019entity} proposed the ESAFN model, which incorporates attention mechanisms, gating mechanisms, and bilinear interactions to capture intra- and inter-modality dynamics for entity-level multimodal sentiment analysis. In 2020, Xu et al. \cite{xu2020social} proposed the AHRM model, which employs a progressive attention module to capture relationships between image and text, ultimately generating a combined image-text representation. 

Additionally, there are BERT-based models,  including TomBERT \cite{yu2019adapting}, which modify the BERT architecture \cite{devlin2018bert} to obtain a target-sensitive visual representation. In 2021, Khan et al. \cite{khan2021exploiting} proposed the EF-CapTrBERT model, which utilizes an object detection transformer to generate an image caption and then construct an auxiliary sentence for multimodal aspect sentiment analysis. In 2022, Yu et al. \cite{yu2022targeted} proposed an Image-Target Matching (ITM) network to obtain image representations based on image-target relevance, thereby enhancing multimodal sentiment analysis through Transformer-based fusion. In 2023, Zhao et al. \cite{zhao2023fusion} proposed the FGSN model, which employs graph convolutional networks on text dependency trees to extract contextual and aspect representations while utilizing positional and channel attention for image features, ultimately fusing these representations for sentiment polarity classification. In 2024, Yang et al. \cite{yang2024macsa} proposed the MGAM model, which constructs an auxiliary sentence and heterogeneous graphs to learn cross-modal interaction.

\subsection{Multimodal Sentiment Analysis in Vietnamese} \label{sec:msa_vn}

The availability of benchmark datasets for multimodal ACSA in low-resource languages, such as Vietnamese, is limited. Currently, multimodal Vietnamese datasets mainly focus on the visual question answering task, including ViVQA \cite{tran2021vivqa}, ViCLEVR \cite{tran2023viclevr}, OpenViVQA \cite{nguyen2023openvivqa}, and EVJVQA \cite{nguyen2023evjvqa} datasets. In the field of sentiment analysis, it is only focused on text-based data \cite{mai2018aspect,ho2020emotion,van2021two, luc2021sa2sl,thanh-etal-2021-span,van2022joint}. Currently, there is no multimodal dataset for sentiment analysis in Vietnamese.

In summary, existing datasets ignore fine-grained annotation in images, which limits the alignment of information between different modalities. Additionally, these datasets only include a single image, which is insufficient for addressing real-world problems. Therefore, we propose a new benchmark dataset and method to address this issue. We present the details in the following sections.

\section{Benchmark Dataset Creation} \label{sec:masc_dataset}

In this section, we describe the multimodal ACSA task and introduce a new benchmark dataset specifically designed for Vietnamese multimodal aspect category sentiment analysis, named ViMACSA. A distinguishing characteristic of the ViMACSA dataset lies in its inclusion of fine-grained annotations for both text and image. Notably, it incorporates fine-grained bounding box annotations, which enhance the analysis capabilities for visual elements within the dataset. The dataset creation process was carried out in three stages: Data Collection (Section \ref{sec:data_collect}), Data Annotation (Section \ref{sec:anno_guide}), and Data Validation (Section \ref{sec:data_validation}). 
% This process is illustrated in Figure {}.

\subsection{Task Definition}
Based on previous research on  aspect-based sentiment analysis \cite{zhang2022survey} and multimodal aspect-based sentiment analysis \cite{zhao2024survey}, MACSA is defined as follows. For each text-image pair, we have a textual content (S) containing n words $S = \{w_1, w_2, ..., w_m\}$ and associated images $I = \{I_1, I_2, ..., I_k\}$. We define a set of aspect categories $A = \{A^1, A^2, ..., A^n\}$. Where m, k, and n denote the number of words in the textual content, the number of images, and the number of predefined aspects, respectively. Given an input pair (S, I) and a specific target aspect $A^n$, the goal is to determine the sentiment label associated with that aspect. $(S, I, A^n)$ to Y, where Y consists of ("none", "negative","neutral" and "positive"). The "none" label implies that aspect $A^n$ is absent from both the textual and visual modalities.

\subsection{Data Collection} \label{sec:data_collect}
To construct a highly authentic and reliable dataset comprising both images and text, we collected user-generated multimodal reviews from Traveloka\footnote{\href{https://www.traveloka.com/vi-vn}{https://www.traveloka.com/vi-vn}}, a Vietnamese tourism website that also provides booking services. The unlabeled dataset consists of 8,000 samples, each comprising a review accompanied by up to 7 images. 

Next, our process involved annotating fine-grained elements for each image by using the object detection tool x-anylabeling\footnote{\href{https://github.com/CVHub520/X-AnyLabeling}{https://github.com/CVHub520/X-AnyLabeling}}. This helped us to detect Regions of Interest (RoIs) automatically. After removing confusing and contradictory data samples through the labeling process, we obtained a dataset of 4,876 text-image pairs that include RoIs. This dataset provides a comprehensive and reliable resource for researchers in the field of Vietnamese multimodal sentiment analysis. It is worth noting that all user reviews are public and do not involve personal privacy.

\subsection{Dataset Annotation Process and Guidelines} \label{sec:anno_guide}

\subsubsection{Aspect Category Definition}

After reviewing previous ABSA research both domestically and internationally \cite{pontiki2016semeval,nguyen2018vlsp,yang2024macsa}, we chose to utilize the Aspect Categories presented in study \cite{yang2024macsa}: Location, Food, Room, Facilities, Service, and Public Area. These aspect categories thoroughly assess hotel-related user concerns and can be applied independently to text or image data.

\subsubsection{Data Annotation} \label{sec:annotation}

\begin{figure}[H]
    \centering
    \includegraphics[width=1.0\linewidth,height=7cm,keepaspectratio]{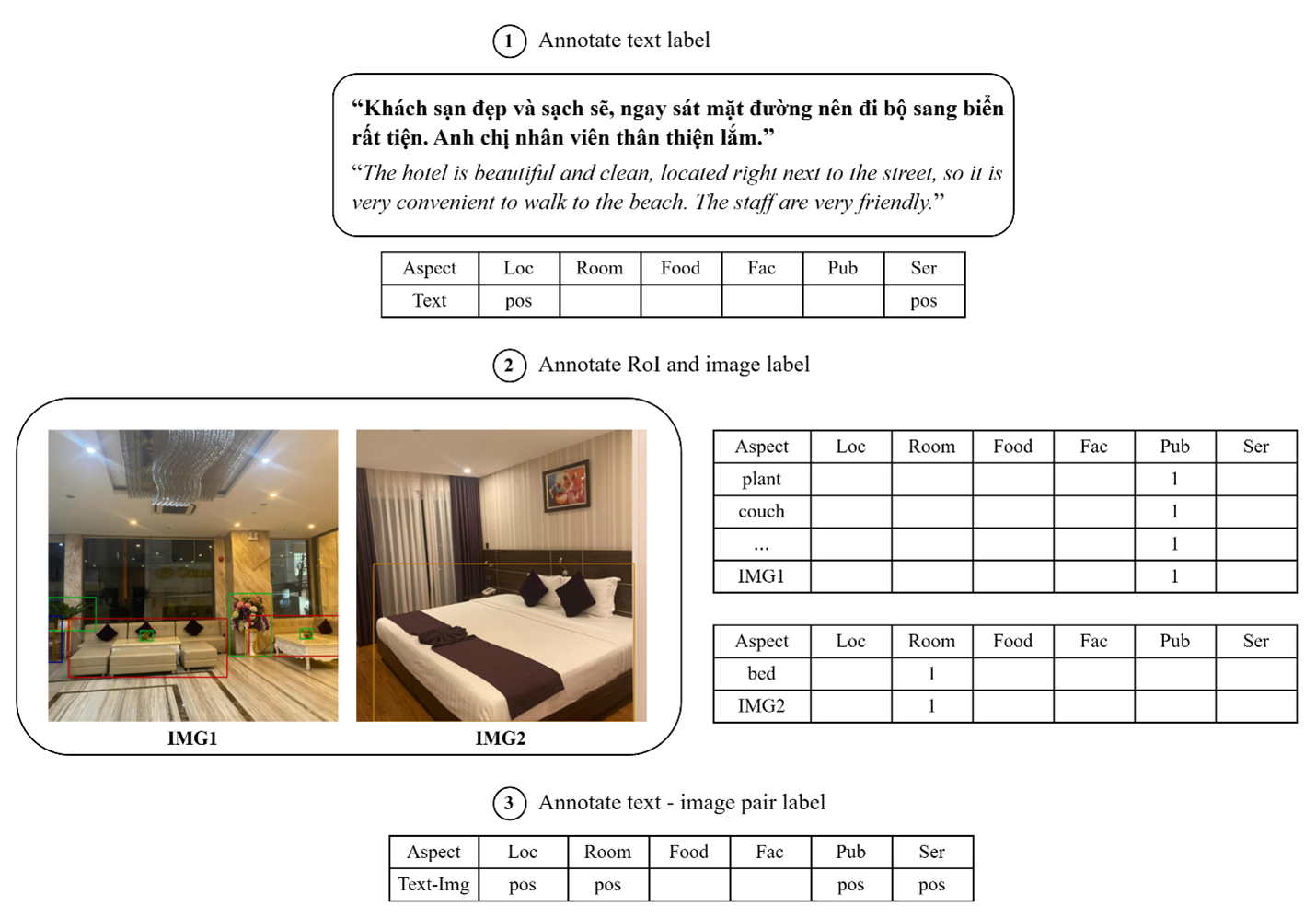}
    \caption{Three-stage annotation process for the ViMACSA dataset.}
    \label{fig:annotate_process}
\end{figure}

The construction process of the ViMACSA dataset in our study comprises a three-stage annotation process: Text Annotation, Image Annotation, and Text-Image Annotation. Figure \ref{fig:annotate_process} illustrates the detailed annotation process for our dataset.

In the first stage of our process, we perform text annotation. This process involves annotating text modality to six predefined aspects and corresponding sentiment polarity. The aspect-sentiment polarity is classified as 0 for irrelevant, 1 for negative, 2 for neutral, and 3 for positive. We used Label Studio\footnote{\href{https://labelstud.io/}{https://labelstud.io/}}, a web-based annotation tool with a user-friendly interface, to accomplish this task easily.

The second stage of our process is called Image Annotation. This stage involves detecting RoIs within an image and annotating specific aspects of each image or RoI. Specifically, we use the YoloV8\footnote{\href{https://github.com/ultralytics/ultralytics}{https://github.com/ultralytics/ultralytics}} model within x-anylabeling, which automatically detects RoIs in each image. Once the RoIs are detected, we remove any irrelevant or incorrect ones and merge similar RoIs. Then, we annotate the image modality and the remaining RoIs to predefined aspects. For example, if the image is a room and the RoI is a bed, then both the image and the RoI will be annotated as "Room". It can be challenging to accurately annotate sentiment labels to images and RoIs due to the difficulty in determining the sentiment conveyed by visual content in the hotel domain \cite{yang2024macsa}. To avoid misleading information between annotators, we do not annotate sentiment labels to our images and RoIs.

In the last phase of our process, we annotate sentiment labels to the text-image pairs for the aspect categories identified in the second stage. If an aspect category already has a sentiment label from the first stage, it will not be changed. However, for new aspect categories that appear in the second stage, we annotate sentiment labels based on the text modality.

In order to maintain the quality of the dataset, a comprehensive annotation guideline has been created. This guideline includes a detailed description of the annotation process, including the definition of aspect category and annotation rules. For more information about the guidelines and the definition of aspect category, please refer to \ref{sec:detail_annotation}.

The annotation guidelines are continuously updated throughout the annotation process to ensure consistency and cover all possible scenarios annotators may encounter. Finally, we obtained the ViMACSA dataset with fine-grained annotations on text and images that were carefully annotated by two annotators. This dataset is valuable for researchers in the fields of natural language processing and multimodal sentiment analysis, particularly in the field of Vietnamese language processing. 

\subsection{Dataset Validation}\label{sec:data_validation}
In order to ensure the quality of annotations and the reliability of our dataset, we have employed a multi-round approach for the annotation process. We use Cohen's Kappa \cite{cohen1960coefficient} and Intersection over Union (IoU) \cite{rezatofighi2019generalized} to estimate inter-annotator agreement, which measures the level of agreement among annotators regarding their labels. Specifically, we calculate Cohen's Kappa scores for labeling tasks related to aspect category and sentiment polarity (Section \ref{sec:annotation}), and we calculate IoU scores for the task of identifying RoI positions in images. 

\begin{figure}[H]
    \centering
    \includegraphics[height=5cm]{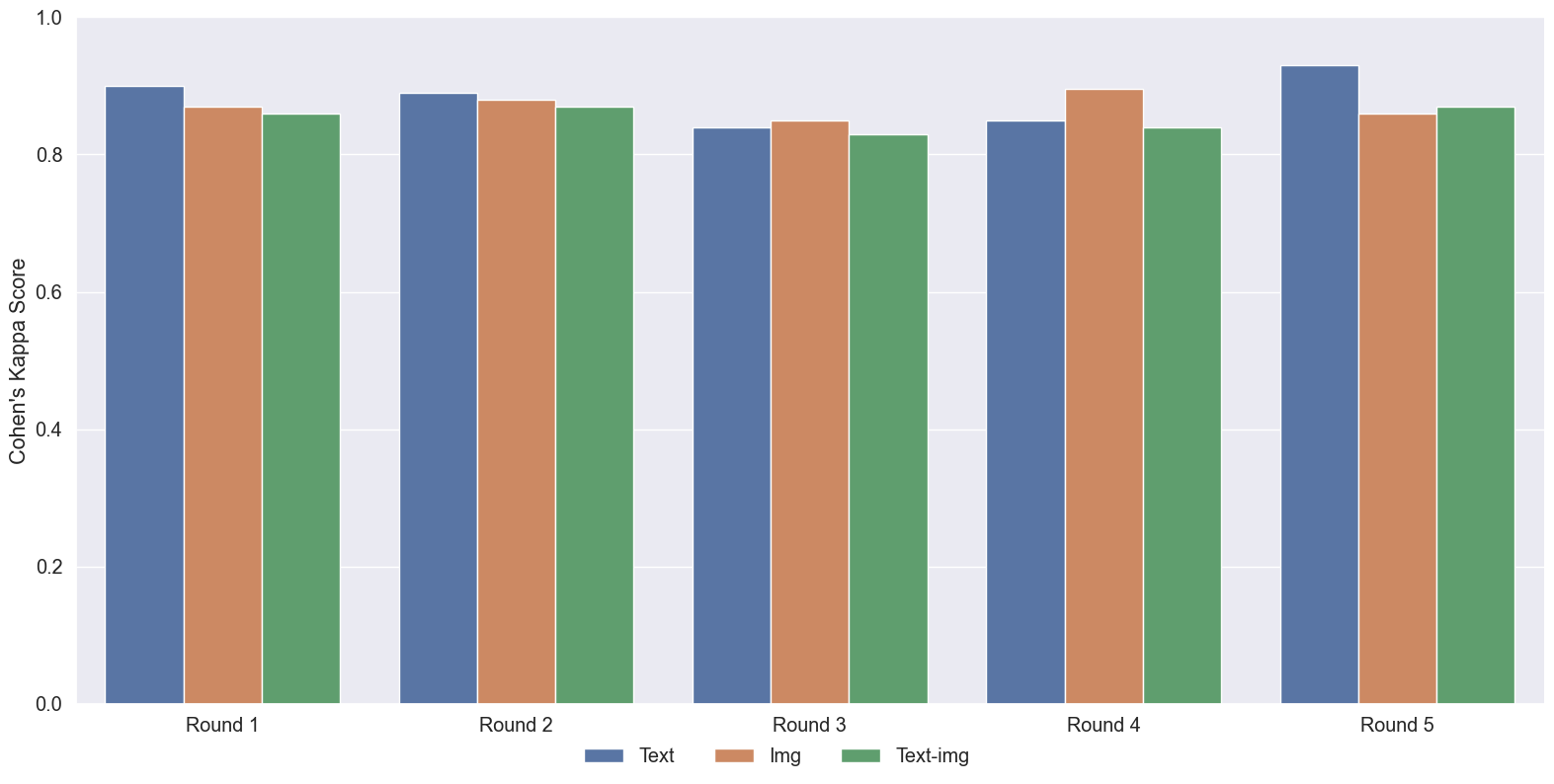}
    \caption{Cohen's Kappa Score of the training phase.}
    \label{fig:fig_aspect_anno}
\end{figure}

\begin{figure}[H]
    \centering
    \includegraphics[height=5cm]{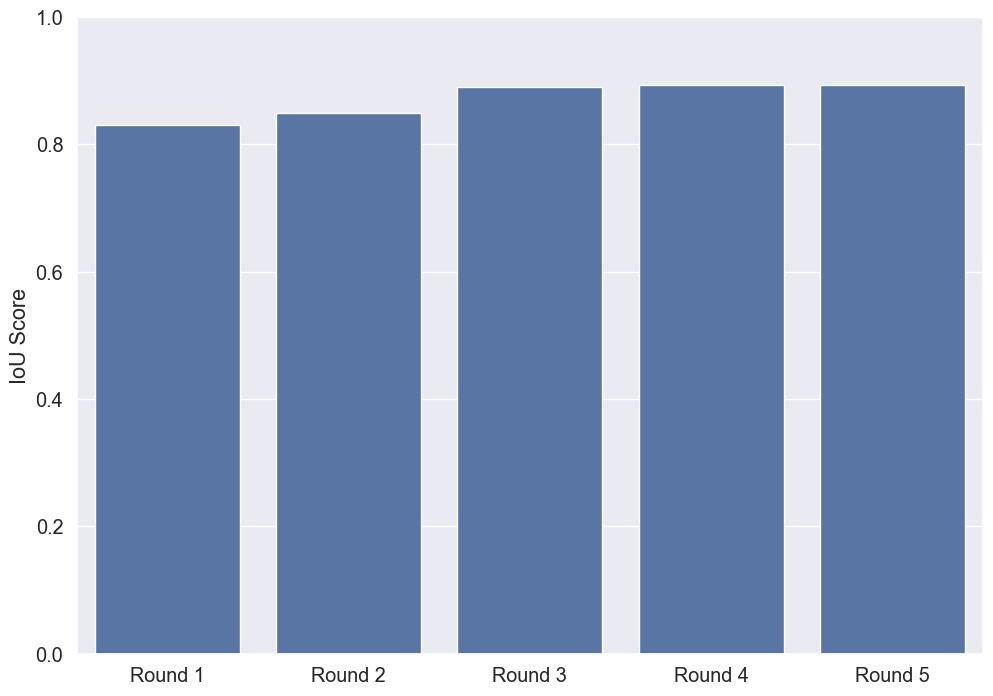}
    \caption{IoU Score of the training phase.}
    \label{fig:fig_iou_anno}
\end{figure}

Our annotation process is divided into two phases: the training phase and the labeling phase. The training phase consists of five rounds, with each round comprising 100 samples. According to \cite{mchugh2012interrater}, the annotation process was finished when the inter-annotator agreement of the annotations reached more than 0.80. As shown in Figure \ref{fig:fig_aspect_anno} and Figure \ref{fig:fig_iou_anno}, the inter-annotator agreement between two annotators is consistently greater than 0.8 in the five rounds.

Once the training phase is complete, we move to the labeling phase, where our annotators annotate the remaining dataset. In cases where samples are challenging, we determine the label based on consensus among annotators.

\subsection{Dataset Analysis}

The statistics of our dataset are presented in Table \ref{tab:stat_overview}. Our dataset consists of 4,876 text-image pairs, which contains fine-grained annotation for both text and image. Upon conducting a thorough analysis, we observed that the aspect category and sentiment labels within the dataset are imbalanced, with the majority of sentiment labels being positive. 

\begin{table}[H]
\centering
\caption{The overview statistics of ViMACSA dataset.}
\label{tab:stat_overview}
\resizebox{\textwidth}{!}{
\begin{tabular}{lcccccccc}
Set   & Review & Average   length & Avg   aspect/review & Positive & Neutral & Negative & Image & RoI  \\ \hline
Train & 2,876   & 42.42            & 3.01                & 6,421     & 1,402    & 830      & 5,428  & 8,656 \\ \hline
Dev   & 1,000   & 39.36            & 2.98                & 2,230     & 463     & 291      & 1,789  & 2,880 \\ \hline
Test  & 1,000   & 42.17            & 2.98                & 2,178     & 485     & 318      & 1,841  & 3,097 \\ \hline
\end{tabular}
}
\end{table}

The number of aspects for text-only modality and text-image pair is shown in Table \ref{tab:stat_modality}. The dataset annotated on both text and image contains 36.51\% more aspects than the text-only modality, demonstrating its potential to address the issue of implicit aspects in user reviews. It indicates that multimodal data holds much information, and analyzing it requires comprehensive consideration of information from both modalities.

\begin{table}[H]
\centering
\caption{Number of Aspects per Modality.}
\label{tab:stat_modality}
\begin{tabular}{lc}
Modality   & Number of aspect \\ \hline
Text-only  & 10,708            \\ \hline
Text-Image & 14,618            \\ \hline
\end{tabular}
\end{table}

\begin{figure}[H]
    \centering
    \includegraphics[width=1.0\linewidth,height=9.5cm,keepaspectratio]{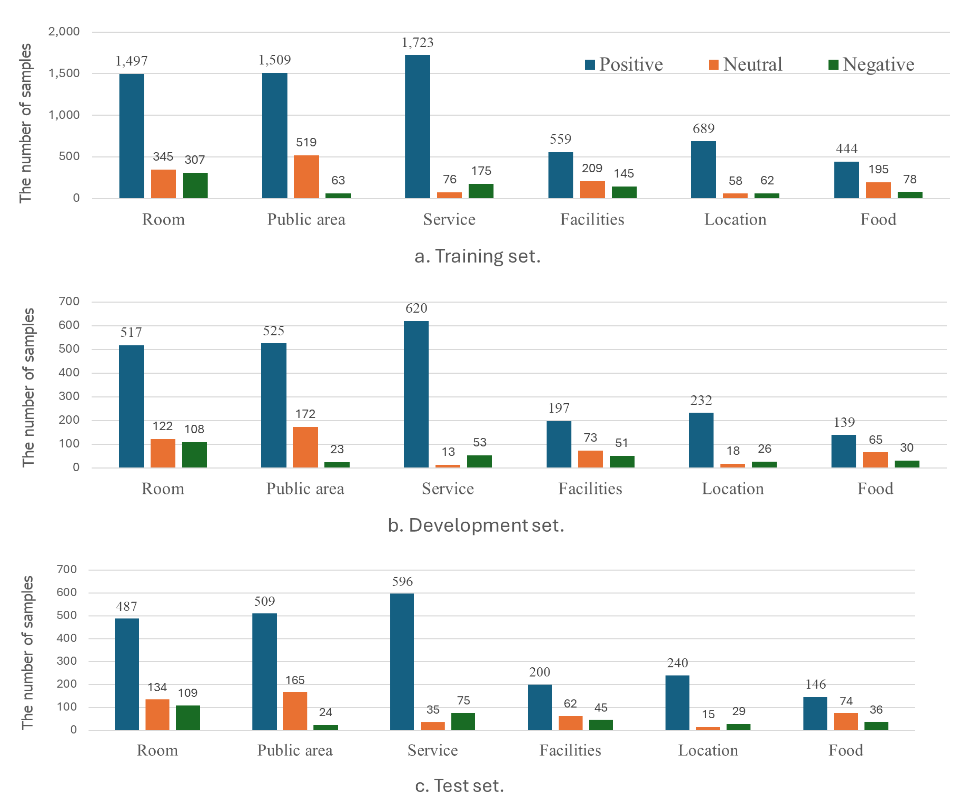}
    \caption{Distribution of 6 Aspect Categories in ViMACSA dataset.}
    \label{fig:aspect_dist}
\end{figure}

Figure \ref{fig:aspect_dist} illustrates the distributions of 6 aspect categories in the dataset. As our study focuses on the hotel domain, aspects such as Room and Service are the most frequently mentioned in user reviews, while the Food aspect has the lowest frequency of mentions.

\section{Our Proposed Framework} \label{sec:proposed_fcmf}
In this section, we describe our proposed FCMF framework in detail.

\subsection{Overview of our framework}

\begin{figure}[H]
    \centering
    \includegraphics[width=1.0\linewidth]{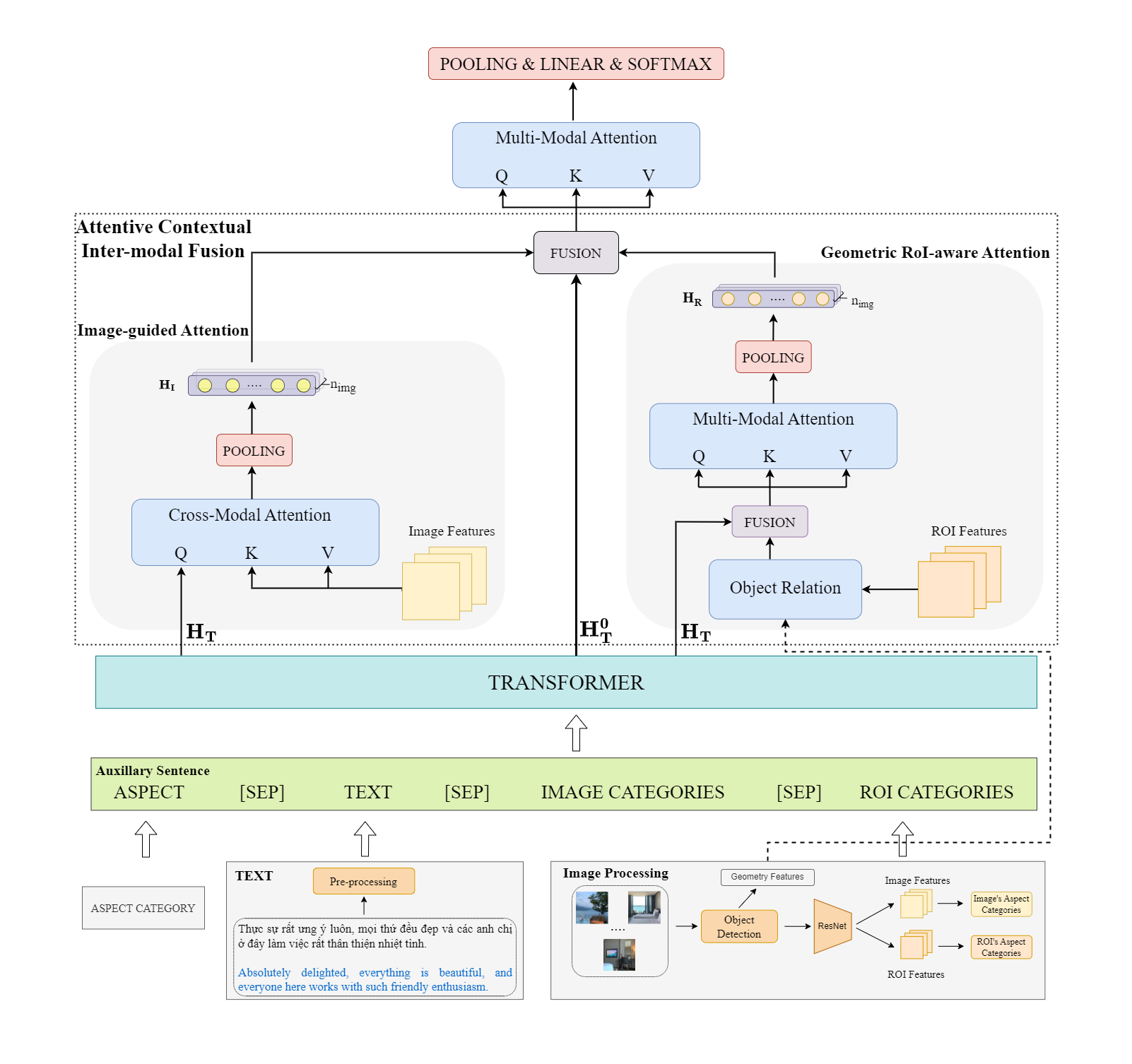}
    \caption{The overview architecture of the proposed FCMF framework.}
    \label{fig:model}
\end{figure}

Our proposed framework consists of four modules: Image Processing, Auxiliary Sentence, Image-guided Attention, and Geometric RoI-aware Attention. The whole architecture is illustrated in Figure \ref{fig:model}.

Given a multimodal input sample $m_i = (A^n, T_i, I)$ comprising an aspect category $A^n$, context $T_i$, and a list of images $I$. For the {\bf textual context}, we initially employ object detection on each image to obtain Regions of Interest (RoIs). We then use a CNN-based model to extract visual features and get the aspect categories $A^I$ and $A^R$ for images and RoIs, respectively. These aspect categories, $A^I$ and $A^R$, are then used to construct an auxiliary sentence derived from the context $T_i$ and the aspect category $A^n$. By utilizing the language model in the sentence classification mode, we feed $(A^n, T_i, A^I, A^R)$ through the language model to generate a textual feature vector. For the {\bf image context}, we utilize Cross-Modal Attention mechanisms to guide the model, ensuring that it assigns high attention weights to image regions that are relevant to the textual context. For the {\bf RoI context}, we employ Bounding Box Relational encoding, which integrates information about spatial relationships between RoIs through Geometric Attention. Subsequently, we utilize Multi-modal Attention to capture the interaction between RoIs and textual context. Finally, we aggregate the textual, image, and RoI representations using Multi-Modal Attention and then feed the multi-modal representation into a softmax function for sentiment classification.

In the following subsections, we introduce each module in detail.

\subsection{Image Processing}
The image processing module encompasses three separate tasks: identifying fine-grained elements, extracting visual features, and detecting aspect categories.

\subsubsection{Aspect Category Detection}
Given a set of images $I=\{I_1,I_2,...,I_K \}$, applying the YOLOv8 algorithm for object detection on each image $I_k$ yields a set of Region of Interest (RoIs) $R^k=\{R_1^k,R_2^k,...,R_J^k \}$ and geometry features (i.e., box coordinates) $G^k=\{G_1^k,G_2^k,...,G_J^k \}$, where K and J denote the number of images and RoIs, respectively. These images serve as input for training a multi-label image classification model to predict image categories. We use one of the state-of-the-art image recognition models called ResNet-152 \cite{he2016deep} models for this task.

Considering that each RoI corresponds to only one aspect category, a multi-class classification model is trained for this purpose. Finally, we obtain a set of image categories $C^I=\{C_1^I,…C_n^I \}$ and RoI category $C^{R^k}$ where $C_n^I \in \{0,1\}$ and $C^{R^k} \in \{0,…,n\}$, indicating whether the image set and RoI set are related to predefined aspects $A^n$.

\begin{equation}
    R^k = Yolov8(I_k), k \in [1,K] 
\end{equation}

\begin{equation}
    C^I = ResNet(I), I = \{I_1, I_2, ... , I_k\} 
\end{equation}

\begin{equation}
    C^{R^k} = ResNet(R^k), R^k = \{R^k_1, R^k_2, ..., R^k_j\}
\end{equation}

\subsubsection{Visual Features Extraction}

For the associated image $I_k$ or RoI $R_j^k$, we start by resizing it to 224 x 224 pixels. We then use ResNet-152 \cite{he2016deep} to extract visual features from both the image and the RoI. Next, we obtain the output of the last convolutional layer for the image, while applying mean-pooling over the spatial dimension for the RoI. Finally, we use a linear function to project the visual features into the same textual feature space, as described in \cite{yu2019adapting}, where $W_I,W_R \in \mathbb{R}^{d \times 2048}$.
\begin{equation}
    ResNet(I_k) = \{r_h \vert r_h \in \mathbb{R}^{2048} ,h=1,2,…,49 \}
\end{equation}

\begin{equation}
    ResNet(R^k_j) = \{r_j \in \mathbb{R}^{2048} \}
\end{equation}

\begin{equation}
    v^{I_k} = W_I ResNet(I_k)
\end{equation}

\begin{equation}
    v^{R_j^k} = W_R ResNet(R_j^k)
\end{equation}

% Once we have obtained the features of each image, we combine them all into a single image feature. The RoI’s feature is as same as the image. Algorithm \ref{alg:alg1} summarizes the main steps for the Image Processing module.

Once we have obtained the features of each image, we combine them into a unified image representation. We also apply the same feature extraction process to the RoI. Algorithm \ref{alg:alg1} provides a step-by-step breakdown of the Image Processing module.

\begin{equation}
    v^I = \{v^{I_k} \vert v^{I_k} \in \mathbb{R}^{d\times49}, k \in [1,K] \}   
\end{equation}

\begin{equation}
    v^{R^k} = \{v^{R_j^k} \vert v^{R_j^k} \in \mathbb{R}^{d}, j \in [1,J] \}   
\end{equation}

\begin{equation}
    v^{R} = \{v^{R^k} \vert v^{R^k} \in \mathbb{R}^{d \times j}, k \in [1,K] \}   
\end{equation}

\begin{algorithm}[H]
\caption{Step-by-step for the Image Processing module.}\label{alg:alg1}
\begin{algorithmic}

\Procedure{ImageProcessing}{images}

\State $\text{image\_features} \gets [], \text{roi\_features} \gets [] $

\State $\text{image\_categories} \gets [], \text{roi\_categories} \gets []$

\State $\text{geometry\_features} \gets []$

\For{i: [1,K]} 

    \State $\text{roi\_list} \gets \text{YoloV8(images[i])}$
    
    \State $\text{image\_i\_category} \gets \text{ResNet-152(images[i])} $
    
    \State $\text{image\_i\_features} \gets \text{image\_encoder(resnet,images[i])}$
    
    \State $\text{roi\_i\_categories} \gets []$
    \State $\text{roi\_i\_features} \gets []$
    
    \State $\text{geometry\_features\_i} \gets []$
    
    \For{j: [1,J]}
        \State $\text{x,y,w,h} \gets \text{roi\_list[j]}$
        \State $\text{roi\_i\_j} \gets \text{get\_roi(x,y,w,h)}$
        \State $\text{roi\_i\_j\_category} \gets \text{ResNet-152(roi\_i\_j)}$
        \State $\text{roi\_i\_j\_features} \gets \text{roi\_encoder(resnet,roi\_i\_j)}$
    
        \State $\text{roi\_i\_categories.append(roi\_i\_j\_category)}$
        \State $\text{roi\_i\_features.append(roi\_i\_j\_features)}$
        \State $\text{geometry\_features\_i.append([x,y,w,h])}$
    \EndFor
    
    \State $\text{image\_features.append(image\_i\_features)}$
    \State $\text{roi\_features.append(roi\_i\_features)}$
    \State $\text{geometry\_features.append(geometry\_features\_i)}$
    \State $\text{roi\_categories.append(roi\_i\_categories)}$
    \State $\text{image\_categories.append(image\_i\_category)}$
\EndFor
\EndProcedure

\end{algorithmic}
\end{algorithm}

\subsection{Auxiliary Sentence}
% XLM-RoBERTa \cite{conneau2019unsupervised}, a variant of BERT \cite{devlin2018bert}, is trained on a large multilingual corpus. It has the ability to derive contextualized word representations and can learn how to align multiple sentences. We use it as the transformer model to extract the textual features for our framework.

XLM-RoBERTa \cite{conneau2019unsupervised}, a variant of BERT \cite{devlin2018bert}, is trained on a large multilingual corpus that supports 100 languages. It has the ability to derive contextualized word representations and can learn how to align multiple sentences. Since our framework aims to be universal, XLM-RoBERTa is a great choice due to its multilingual capabilities, which allow us to extract rich textual features across multiple languages, including Vietnamese. This ability to process and understand text in diverse languages makes XLM-RoBERTa the ideal backbone for our framework's text feature extraction.

The construction of the Auxiliary Sentence has achieved remarkable results \cite{sun2019utilizing, khan2021exploiting, yang2024macsa}. Inspired by this, we employ XLM-RoBERTa for the Auxiliary Sentence task, which transforms each sentence S into four sub-sentences: aspect category $A^n$, context T, image's categories $A^I$, and RoI's categories $A^R$ and concatenate them to form the input sequence for XLM-RoBERTa: "<s> $A^n$ </s></s> T </s></s> $A^I$ </s></s> $A^R$ </s>". Then, the XLM-RoBERTa embedding layer transforms the input sequence into a textual embedding sequence ${\bf X} = ( x_1, x_2, …, x_N )$, where $x_i \in \mathbb{R}^d $ is the input representation, and N is the maximum length of the sequence. 

This textual embedding sequence ${\bf X}$ fuses visual information into the text modality's feature space, enriching the textual representation with semantic information derived from the visual modality. Finally, we feed ${\bf X}$ into the XLM-RoBERTa encoder to obtain the final hidden state of the first <s> token.

\begin{equation}
    {\bf H_T} = {\text{XLM-RoBERTa}} ({\bf X})
\end{equation}

\begin{equation}
    {\bf H_{<s>}} = {\bf H_T^0}
\end{equation}
\subsection{Image-guided Attention}
Our Image-guided Attention module is inspired by tomBERT \cite{yu2019adapting}, but in contrast to tomBERT, we use input sequence instead of aspect category to capture context-image relevance and help focus on sentiment words within context.

To achieve this goal, we apply Cross-modal Attention \cite{tsai2019multimodal} to model the interaction between context and image, which treats the hidden states of the input sequence ${\bf H_T}$ as a query, and the image features ${  v^{I_k}}$ as keys and values, as follows:

\begin{equation}
    {\bf H_{{I_k}}} = {\text {CM-Attention}}({\bf H_T}, { v^{I_k}}, { v^{I_k}})
\end{equation}

where ${\bf H_{{I_k}}} \in \mathbb{R}^{d \times N}$  is the generated image-based context representation. Then, we obtain the representation of the first token (i.e., ${\bf H^0_{I_k}}$ ), which is a fused representation with textual representations as the query. Finally, we concatenate all of these to produce a unified image-based context representation:

\begin{equation}
    {\bf H_{I}} = \{ {\bf H^0_{{I_1}}}, {\bf H^0_{{I_2}}}, ..., {\bf H^0_{{I_k}}} \}
\end{equation}

\subsection{Geometric RoI-aware Attention} 
Intuitively, the spatial relationships among objects play a crucial role in understanding the content within an image and have been shown to improve the model performance \cite{hu2018relation,herdade2019image}. This can help identify an image's aspect category, connect different objects, and provide additional information. For example, if a sofa and a coffee table are placed close to each other and surrounded by small plants, the image may depict the lobby of a hotel.

To accomplish this objective, we feed geometry features and RoIs features into the Object Relation module to automatically model the interactions and learn the geometric relationships between different RoIs.

\begin{equation}
    {\bf H_O^{I_k}} = {\text{Object-Relation} (v^{R^k}, G^k)} 
\end{equation}

where ${\bf H_O^{I_k}} \in \mathbb{R}^{d \times J}$, where J denote the number of RoI. Then, we concatenate ${\bf H_O^{I_k}}$ and ${\bf H_T}$ together as: ${\bf H_{T\&O}^{I_k}} = {\bf H_T} \oplus {\bf H_O^{I_k}}$ and feed them to Multi-Modal Attention for multimodal fusion:

\begin{equation}
    {\bf H^{{I_k}}_{T\&R}} = {\text {MM-Attention}}({\bf H_{T\&O}^{I_k}}, {\bf H_{T\&O}^{I_k}}, {\bf H_{T\&O}^{I_k}})
\end{equation}

where ${\bf H^{{I_k}}_{T\&R}} \in \mathbb{R}^{d \times (N+J)}$ is the generated spatial relative RoI-based context representation. Similar to Image-guided Attention, we obtain the representation of the first token. We then combine these representations to create a contextual representation based on the relative position of RoI:

% \begin{equation}
%     {\bf H^{I_k}_{R}} = {\text {Avg}}(h^{R_0},...,h^{R_j})
% \end{equation}

\begin{equation}
    {\bf H_{R}}= \{ {\bf H^{I_1,0}_{T\&R}}, {\bf H^{I_2,0}_{T\&R}}, ..., {\bf H^{I_k,0}_{T\&R}} \}
\end{equation}

\subsection{Sentiment Detection}
In the final stage, we concatenate the text features ${\bf H_{<s>}}$, image features ${\bf H_{I}}$ and the RoI features ${\bf H_{R}}$ to form a unified multimodal representation. This representation is then passed to another Multimodal Attention layer to capture the dependencies and interactions among the different modalities:

\begin{equation}
    {\bf H^M} =  {\bf H_{<s>}} \oplus {\bf H_{I}} \oplus {\bf H_{R}} 
\end{equation}

\begin{equation}
    {\bf H} = {\text{MM-Attention}}({\bf H^M},{\bf H^M},{\bf H^M})  
\end{equation}

After obtaining the final hidden state of the first token ${\bf H}$, we pass it through a linear function followed by a softmax function for sentiment classification:

\begin{equation}
    P(y) = {\text{ softmax} ({\bf W^T H^0+b})}
\end{equation}

% where ${\bf W^T}$ and ${\bf b}$ are the learnable parameters. To optimize all the parameters in our framework, the objective is to minimize the standard cross-entropy loss function, where D is the number of samples, and N is the number of predefined aspects. The core training steps for the MFAM framework are summarized in Algorithm \ref{alg:alg2}.

where ${\bf W^T}$ and ${\bf b}$ are the learnable parameters. We optimize all parameters in our framework by minimizing the standard cross-entropy loss function, where D is the number of samples, and N is the number of predefined aspects. The core training steps for the FCMF framework are summarized in Algorithm \ref{alg:alg2}.

\begin{equation}
    \mathcal{L} = -\frac{1}{D} \frac{1}{N} \sum_{i=1}^D \sum_{j=1}^N logP(y^{i,j})
\end{equation}

% TODO: Tranning of MFAM
\begin{algorithm}[h]
\caption{The training of FCMF framework.}\label{alg:alg2}
\begin{algorithmic}

\Procedure{Train}{aspect category, text, image\_features, roi\_features, geometry\_features, roi\_categories, image\_categories}

\State $\text{auxiliary\_sentence} \gets \text{construct\_auxiliary(text, image\_categories, roi\_categories)}$

\State $\mathbf{H_T} \gets \text{Transformer(auxiliary\_sentence)}$

\State $\mathbf{H_I} \gets [], \mathbf{H_R} \gets []$

\For{ i: [1,K]}
    \State $\text{h\_i} \gets \text{image\_guided\_attention($\mathbf{H_T}$, image\_features[i])}$
    
    \State $\text{r\_i} \gets \text{geometric\_roi\_attention($\mathbf{H_T}$, roi\_features[i], geometry\_features[i])}$
    
    \State $\text{$\mathbf{H_I}$.append(h\_i)}$
    \State $\text{$\mathbf{H_R}$.append(r\_i)}$
\EndFor

\State $\mathbf{H^M} \gets \mathbf{H_T^0} \oplus \mathbf{H_I} \oplus \mathbf{H_R}$

\State $\mathbf{H} \gets \text{MM-attention($\mathbf{H^M}$)}$

\State $\hat{y} \gets \mathbf{H^0}$

\State \Return $\hat{y}$

\EndProcedure
\end{algorithmic}
\end{algorithm}

\section{EXPERIMENTS AND RESULTS} \label{sec:experiment}
\subsection{Baseline Models}

In this section, we evaluate our framework against several baseline models on our ViMACSA dataset. Several models, marked with *, were initially developed for single-image tasks. We randomly select one image from related images to adapt them for comparison in the Multimodal ACSA task. To comprehensively evaluate our framework, we categorize our experiments into two approaches: Text-Based Models (Section \ref{sec:text_based}) and Text-Image Models (Section \ref{sec:text_image_models}).

\subsubsection{Text-Based Models} \label{sec:text_based}
% Todo: why choose? j
We selected the following text-based models due to their strengths in aspect-based sentiment analysis, offering diverse techniques such as attention mechanisms, convolutional networks, and specialized gating units:

\textbf{Deep Memory Network (MemNet)} \cite{tai2017memnet}: is a deep memory network that applies multi-hop attention between context and aspect. It uses attention mechanisms to capture the relationship between context words and a given aspect category.

\textbf{Gated Convolutional network with Aspect Embedding (GCAE)} \cite{xue2018aspect}: uses specialized gating mechanisms and convolutional neural networks. It employs Gated Tanh-ReLU Units (GTRU) to derive sentiment features based on the given aspect category, and the computations can be easily paralleled during training. 

\textbf{Interactive Attention Networks (IAN)} \cite{ma2017interactive}: utilizes two LSTM networks to independently model the input sentence and aspect category. It then learns the coarse-grained attentions between the context and aspect, and concatenates the resulting vectors for prediction.

\textbf{Local Context Focus Mechanism with RoBERTa (LCF-RoBERTa)} \cite{zeng2019lcf}: utilizes the Local Context Focus (LCF) mechanism. This mechanism incorporates Context Features Dynamic Weighted (CDW) and Context Features Dynamic Mask (CDM) layers to focus on local context words. The LCF design is also augmented with a RoBERTa layer, enabling it to capture long-term dependencies within both local and global contexts.

\subsubsection{Text-Image Models} \label{sec:text_image_models}
% Todo: why choose?

In order to demonstrate the potential and reliability of our proposed framework, we selected the following state-of-the-art text-image models due to their capabilities to handle the complexities of multimodal aspect-based sentiment analysis:

\textbf{Entity-sensitive Attention and Fusion Network (ESAFN)}* \cite{yu2019entity}: aims to capture inter-modality and intra-modality interactions. It employs separate LSTMs for the left context, right context, and target entity. Attention mechanisms refine context representations, and a specialized visual attention mechanism extracts relevant visual components while a gating mechanism mitigates visual noise.

\textbf{Multi-Interactive Memory Network (MIMN)} \cite{xu2019multi}: employs interactive memory networks to model self-modality and cross-modality interactions, focusing on how a given aspect category influences both textual and visual elements.

\textbf{MACSA-BiLSTM} \cite{yang2024macsa}: concatenates aspect category, context, image category, and RoI category to form an input sequence. It then employs the Multi-Interactive Memory Network to learn a fused multimodal representation.

\textbf{Multimodal RoBERTa (mRoBERTa)}* \cite{yu2019adapting}: concatenates the final hidden states of the input sequence with the image features. Then, additional XLM-RoBERTa layers are added to model the interactions between textual and visual representations.

\textbf{Target-Oriented Multimodal RoBERTa (TomRoBERTa)}* \cite{yu2019adapting}: models interactions between textual and visual representation, emphasizing their relationship to a given aspect category. It also utilizes a Target-Image (TI) matching layer to extract relevant visual information to the aspect category.

\textbf{EF-CapTrRoBERTa}* \cite{khan2021exploiting}: utilizes a transformer architecture designed for object detection to generate image captions. These captions are then fed into the XLM-RoBERTa language model by constructing an auxiliary sentence. Finally, the resulting encoded representation is used for multimodal aspect sentiment analysis.

\subsection{Evaluation Metrics}

In this section, we explain the evaluation metrics used in this article. Average macro Precision, Recall, and F1-score are standard metrics used in various classification tasks \cite{khan2021exploiting,yang2024macsa}. However, our ViMACSA dataset has an imbalance in aspect category and sentiment polarity. Therefore, the most suitable metric for this task is the average macro F1-score, which harmonizes Precision and Recall \cite{sokolova2009systematic}. Hence, to evaluate the performance of our FCMF framework, we chose to use the average macro F1-score as the primary metric, while Precision and Recall provide additional information.

\begin{equation}
    Precision_M = \frac{1}{N}\sum_{i=1}^N\frac{TP_i}{TP_i + FP_i}
\end{equation}

\begin{equation}
    Recall_M = \frac{1}{N}\sum_{i=1}^N\frac{TP_i}{TP_i + FN_i}
\end{equation}

\begin{equation}
    F1 = \frac{1}{N}\sum_{i=1}^N\frac{2TP_i}{2TP_i + FN_i + FP_i}
\end{equation}

where M represents macro-averaging, N is the number of predefined aspects, and $TP_i$, $FP_i$, and $FN_i$ denote true positive, false positive, and false negative, respectively.

\subsection{Experimental Settings}

We perform data preprocessing steps to address common issues in the Vietnamese language, such as diacritics and tone marks, by using underthesea\footnote{\href{https://github.com/undertheseanlp/underthesea}{https://github.com/undertheseanlp/underthesea}}. Additionally, we also use word clouds to show the differences in vocabulary between tokenized and untokenized text in \ref{sec:vocal_wl}. These preprocessing steps are essential to ensure the model's performance.

For our FCMF framework, we adapt $\text{XLM-RoBERTa}_{base}$ as the transformer model and utilize YOLOv8 for object detection and the ResNet-152 \cite{he2016deep} for visual features extraction. Our image processing module incorporates seven images, each containing four RoIs. For XLM-RoBERTa-based models, we utilize the Adam optimizer with the learning rate as 3e-5, the number of attention heads as $m=12$, and the dropout rate as 0.1. The batch size is set to 4, and the maximum sentence length input is 170. We use the same five random seeds across all experiments to ensure a fair comparison. To provide a comprehensive evaluation, we report the average performance over five times. All models are implemented using PyTorch with NVIDIA P100 GPU.

\subsection{Experimental Results}

Based on the results presented in Table \ref{tab:main}, we can conclude that the incorporation of additional image information leads to better performance compared to models that only use text data. This implies that images can provide additional information and serve as a supportive factor in the text.

Our proposed FCMF framework demonstrates significant potential in enhancing the multimodal ACSA task. By using all images and fine-grained information between text and images, our framework outperforms other multimodal baseline models, achieving the highest macro-F1 score of 79.73\%.

\begin{figure}[H]
    \centering
    \includegraphics[width=1.0\linewidth,height=7cm,keepaspectratio]{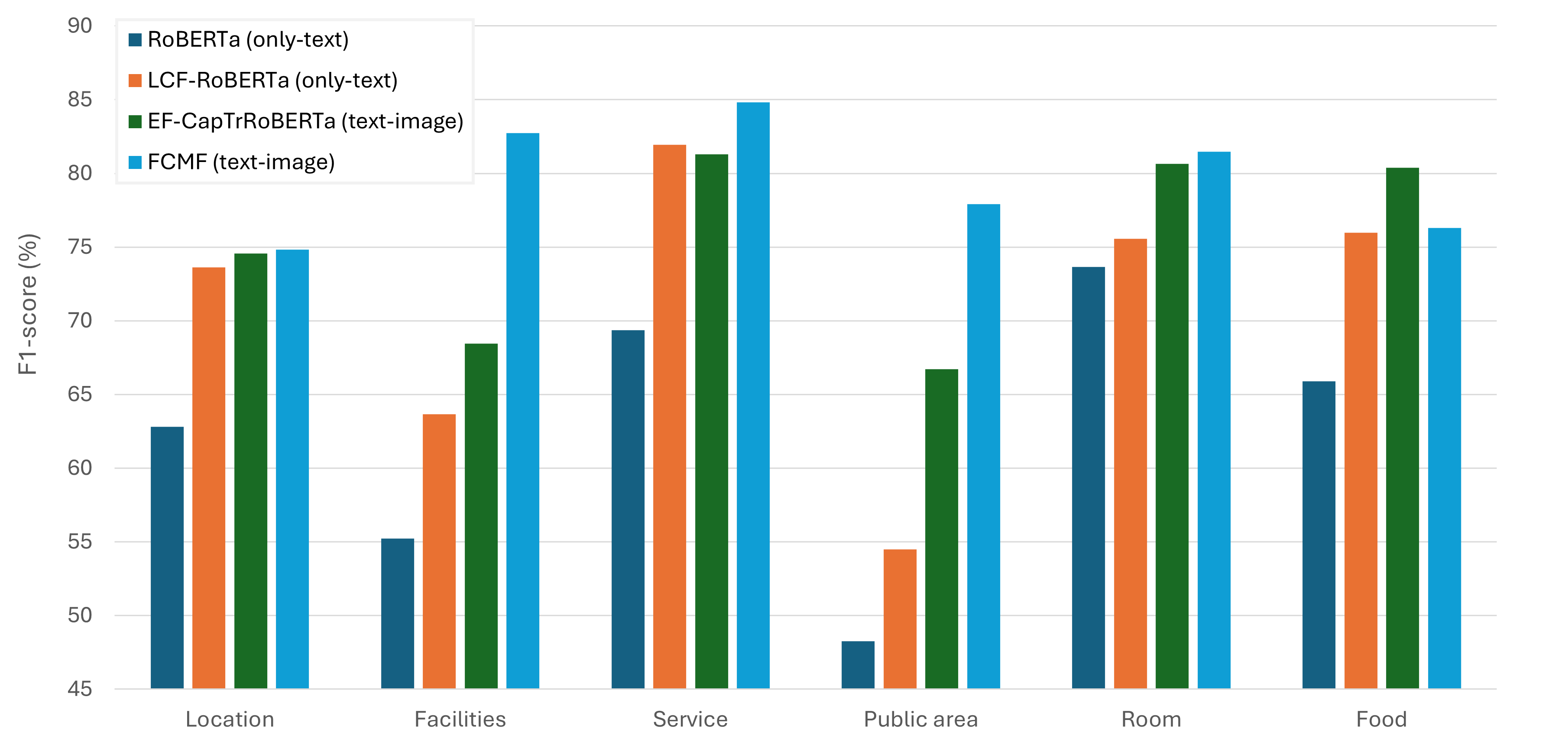}
    \caption{Performance Comparison Between the FCMF Framework with Others in terms of Each Aspect Category.}
    \label{fig:cp_aspect_sentiment}
\end{figure}

% note: ket qua co anh thi tot
Our FCMF framework achieves remarkable results in aspect sentiment polarity prediction, achieving the highest F1 scores across most aspect categories, as shown in Figure \ref{fig:cp_aspect_sentiment}. We also observed that models using both text and image data have higher F1 scores than models using only text data. This indicates that the fine-grained information extracted from images can effectively address the implicit aspect problem. Additionally, it highlights the significance of fine-grained information between text and image in enhancing the model's performance in the MACSA task. 

{\bf Image/RoI Category Detection Result}: We use the ResNet-152 model to train for image/RoI aspect category detection. Experimental results demonstrate that the model achieves an accuracy of 93.97\% for the image category detection task and 82.54\% for the RoI category detection task.

\begin{table}[H]
\centering
\caption{Experiment results on the ViMACSA dataset.}
\label{tab:main}
%\resizebox{\textwidth}{!}{%
% \usepackage{multirow}
\centering
\begin{tabular}{llccc}
\hline
Modality                       & \multicolumn{1}{l}{Models} & Precision & Recall & F1-score \\ \hline
\multirow{3}{*}{Text}          & MemNet                     & 57.71     & 51.53  & 51.88  \\
                               & GCAE                       & 61.92     & 54.06  & 55.39  \\
                               & IAN                        & 65.02     & 59.05  & 60.28  \\ \hline
\multirow{3}{*}{Text + Visual} & ESAFN                      & 64.70     & 59.79  & 60.32  \\
                               & MIMN                       & 67.01     & 62.74  & 63.73  \\
                               & MACSA-LSTM                 & 72.53     & 69.04  & 69.54  \\ \hline
\multirow{2}{*}{Text}          & RoBERTa                    & 66.76     & 61.69  & 62.53  \\
                               & LCF-RoBERTa               & 73.63     & 69.69  & 70.96  \\ \hline
\multirow{4}{*}{Text + Visual} & mRoBERTa                   & 75.55     & 72.08  & 73.24  \\
                               & tomRoBERTa                 & 76.36     & 73.33  & 74.32  \\
                               & EF-CapTrRoBERTa           & 76.74     & 74.68  & 75.40  \\
                               & {\bf FCMF}                       & {\bf 81.43}     & {\bf 78.80}  & {\bf 79.73}  \\ \hline
\end{tabular}

\end{table}

% \begin{table}[H]
% \centering
% \caption{Result per class for only aspect label on FCMF framework.}
% \label{tab:aspect-table}
% \begin{tabular}{lccc}
% Aspect       & Precision & Recall & F1-score \\ \hline
% Location     & 95.51     & 95.57  & 95.53  \\ \hline
% Facilities   & 96.90     & 97.64  & 97.26  \\ \hline
% Service      & 96.26     & 95.81  & 96.02  \\ \hline
% Public area & 94.39     & 95.11  & 94.73  \\ \hline
% Room         & 94.96     & 95.16  & 95.05  \\ \hline
% Food         & 94.98     & 94.63  & 94.78  \\ \hline
% \end{tabular}
% \end{table}

% just focus on aspect absent problem.
% \begin{table}[H]
% \centering
% \caption{F1-score per class for aspect\#polarity label.}
% \label{tab:aspect-polar-table}
% \begin{tabular}{lccc}
%              & Negative & Neutral & Positive \\ \hline
% Location     & 77.19    & 32.65   & 91.89    \\ \hline
% Facilities   & 72.39    & 69.07   & 90.54    \\ \hline
% Service      & 84.73    & 67.88   & 93.07    \\ \hline
% Public\_area & 77.06    & 55.90   & 83.52    \\ \hline
% Room         & 81.94    & 54.74   & 96.79    \\ \hline
% Food         & 47.19    & 68.06   & 90.35    \\ \hline
% \end{tabular}
% \end{table}

\subsection{Impact of Number of Images}

\begin{figure}[H]
    \centering
    \includegraphics[width=0.8\linewidth,height=10cm,keepaspectratio]{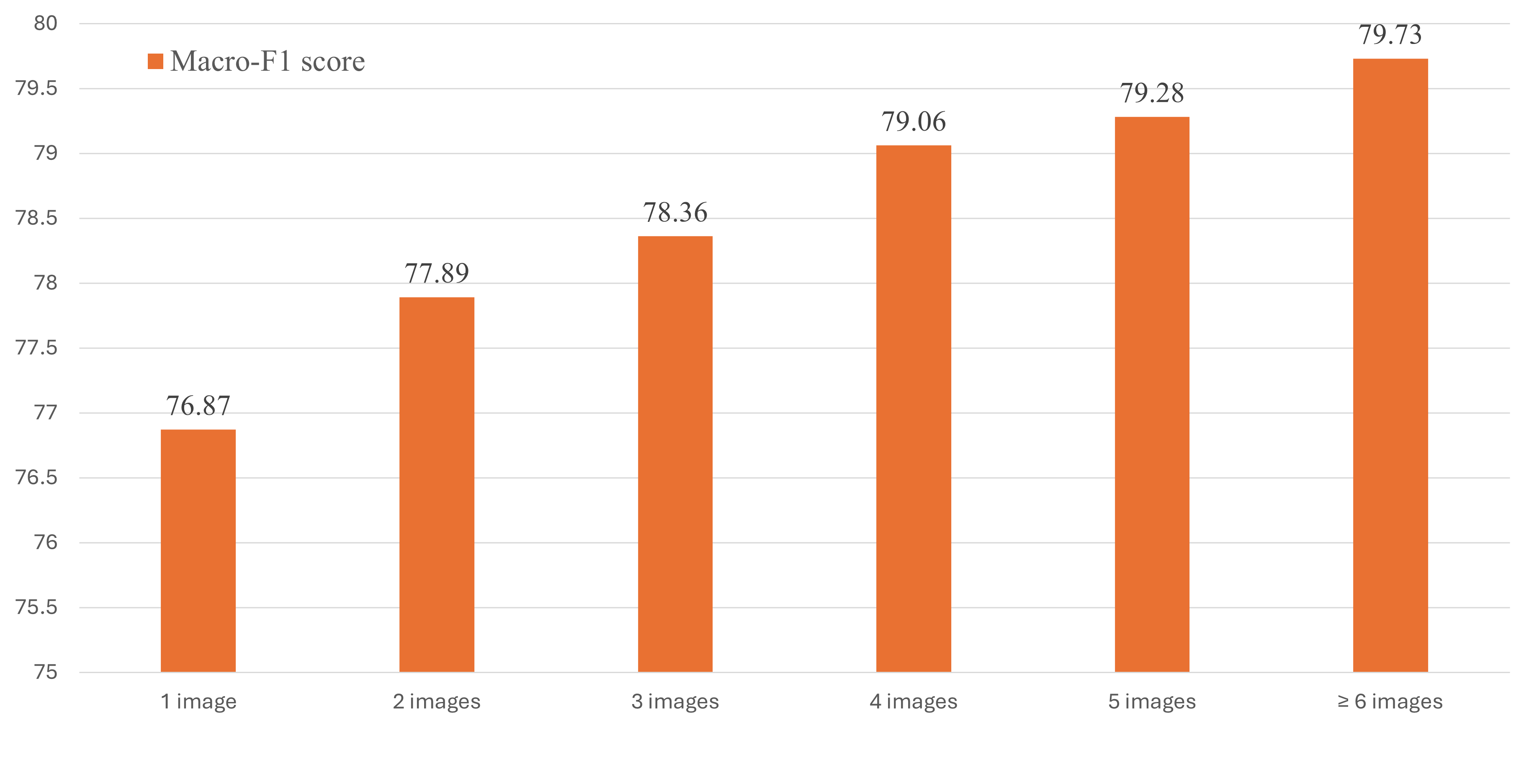}
    \caption{Impact of number of images on FCMF framework.}
    \label{fig:w_img}
\end{figure}
% # note >=6
% Figure \ref{fig:w_img} illustrates the results obtained from our experiments with seven different images. Due to limitations in data collection, the experiment was limited to 7 images

To comprehensively investigate the importance of images, we conducted experiments using varying numbers of images, as shown in Figure \ref{fig:w_img}. Interestingly, using just one image yielded pretty good results, outperforming other baseline models. This demonstrates the effectiveness of using fine-grained images to improve the quality of our framework. Furthermore, we observed that the results improved further as the number of images increased (2.86\% improvement from 1 image to more than 5 images). This claims that our framework can effectively handle multiple images.

\subsection{Ablation study}

% 77.90 77.11 77.09
\begin{table}[H]
\centering
\caption{Ablation study of our proposed FCMF framework.}
\label{tab:ablation}
\begin{tabular}{lccl}
                         & Precision & Recall & F1-score \\ \hline
FCMF                     & 81.43     & 78.80  & 79.73  \\ \hline
- w/o Text Pre-processing &  79.57 & 79.01 & 79.02 (-0.71) \\ \hline
- w/o Geometric RoI-aware Attention & 79.00     & 77.27  & 77.81 (-1.92)  \\ 
- w/o Visual Features & 77.90 & 77.11 & 77.09 (-2.64) \\ 
- w/o Auxiliary Sentence & 65.92     & 65.80  & 59.51 (-20.22) \\ \hline
\end{tabular}
\end{table}

We conduct an experiment to investigate the impact of data preprocessing on the performance of the FCMF framework, as shown in Table \ref{tab:ablation}. Specifically, we remove the preprocessing step in the text and find that this caused our framework to lose significant information about the unique characteristics of the Vietnamese language, such as diacritics, abbreviations, and special characters commonly used on social media\footnote{Refer to \ref{sec:char_data} for further details on the characteristics of social media data and the preprocessing steps used in our ViMACSA dataset.}. As a result, the F1 score decreased by 0.71\%.

Furthermore, we evaluate the impact of each module in our framework. Specifically, removing the Geometric RoI-aware Attention module results in a lack of information about geometric relationships and fine-grained information between different objects, leading to a performance drop of around 1.92 percent points on the F1 score. Furthermore, we find that the Auxiliary Sentence module plays a crucial role in aligning information between image and text (i.e., aspect category). Removing this module significantly decreases performance, around 20 percent points on the F1 score. Finally, we observe that removing the Visual Features Extraction module results in a lack of information on both images and RoIs, which negatively impacts the performance of the FCMF framework. Nevertheless, using the Auxiliary Sentence module helps compensate for this information gap, leading to only a 2.64\% performance drop on the F1 score.

\subsection{Error Analysis}
\label{apx:definition_error}

To compare the predictions of different models, we select a set of representative test samples, as shown in Table \ref{tab:case-study}. In the first row, we observe that other models have incorrectly predicted the sentiment of Service and Facilities\footnote{Refer to \ref{sec:visual_img_rv} for more details on why the FCMF framework makes the correct predictions}. With the help of image and fine-grained elements, our proposed framework can accurately detect the correct aspect category "Service" by interpreting the first image's fine-grained element "staff". Next, in the second row, our framework can focus more on the text and yield results in the correct aspect category, "Food". Moreover, our framework can also capture the context-image relevance, which enables it to correct the wrong predictions made by EF-CapTrRoBERTa and LCF-RoBERTa in predicting "Room" and "Facilities".

\begin{table}[H]
\centering
\caption{Predictions of FCMF, EF-CapTrRoBERTa, and LCF-RoBERTa on several test samples. \text{\cmark} and \text{\xmark}  denote correct and incorrect predictions.}
\label{tab:case-study}
\resizebox{\columnwidth}{!}{
\begin{tabular}{|p{8cm}|c|c|c|c|}
\hline
Input & Human Label & FCMF & EF-CapTrRoBERTa & LCF-RoBERTa \\ \hline

Cá nhân mình thấy rất là okiela , khách sạn gần biển , gần các quán ăn , đi lại rất tiện, mọi thứ đều okie, phòng ốc thế cũng quá okie rồi vì đi chơi suốt nên không cần gì hơn. Nếu tới lần sau vẫn chọn lại khách sạn Soco nhé.

&       % ========== Human label ==========
\multirow{3}{*}{\begin{tabular}[c]{@{}l@{}} \\ \\ \\ \\ \\ \\Location: Positive\\ Room: Positive\\ Facilites: Positive\\ Service: Positive\\ Public area: Positive\end{tabular}}     

&       % ========== FCMF ==========
\multirow{3}{*}{\begin{tabular}[c]{@{}l@{}} \\ \\ \\ \\ \\ \\Location: \textcolor{red}{Positive} \cmark \\ Room: \textcolor{red}{Positive} \cmark \\ Facilites: \textcolor{red}{Positive} \cmark \\ Service: \textcolor{red}{Positive} \cmark \\ Public area: \textcolor{red}{Positive} \cmark \end{tabular}} 
  
&       % ========== EF-CapTrRoBERTa ========== 
\multirow{3}{*}{\begin{tabular}[c]{@{}l@{}} \\ \\ \\ \\ \\ \\Location: \textcolor{red}{Positive} \cmark\\ Room: \textcolor{red}{Positive} \cmark\\ Facilites: None \xmark\\ Service: None \xmark\\ Public area: \textcolor{red}{Positive} \cmark\end{tabular}} 

&       % ========== LCF-RoBERTa ==========
\multirow{3}{*}{\begin{tabular}[c]{@{}l@{}} \\ \\ \\ \\ \\ \\Location: \textcolor{red}{Positive} \cmark \\ Room: \textcolor{red}{Positive} \cmark \\ Facilites: None \xmark\\ Service: None \xmark\\ Public area: None \xmark\end{tabular}} \\

\textit{\textcolor{blue}{I am delighted with my stay at the Soco Hotel. It is conveniently located close to the beach, restaurants, and transportation. The rooms are more than acceptable for my needs, as I spend most of my time outdoors. If I visit again, I will stay at the Soco Hotel. }}

&  &  &  &  \\
\includegraphics[width=0.32\linewidth]{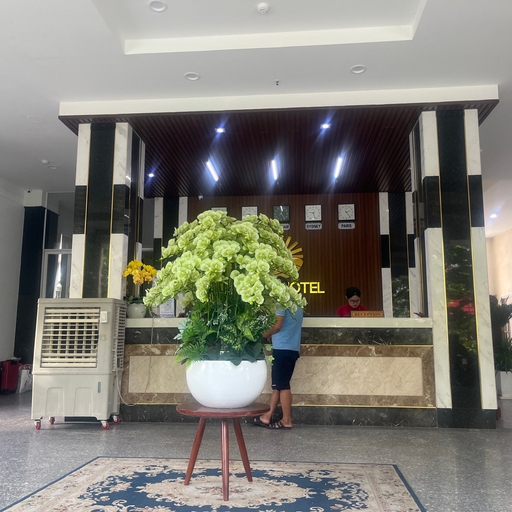} \includegraphics[width=0.32\linewidth]{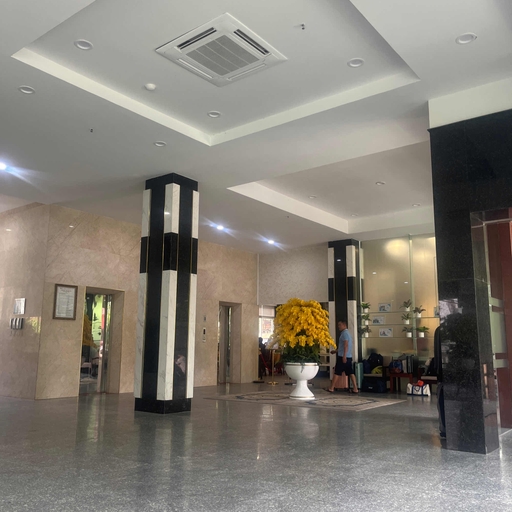}
\includegraphics[width=0.32\linewidth]{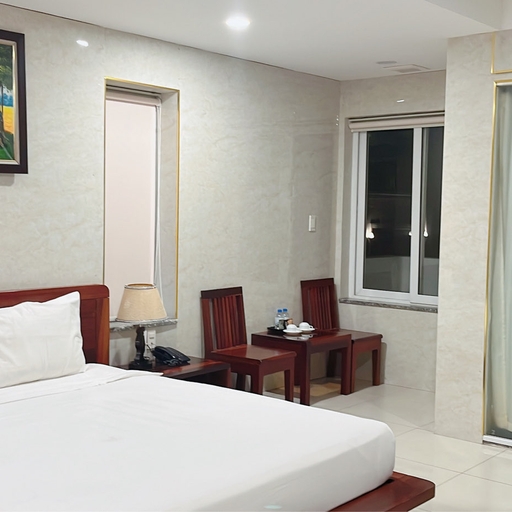} 

\includegraphics[width=0.32\linewidth]{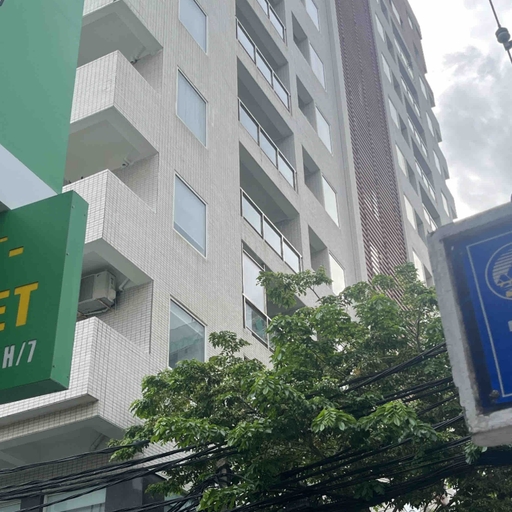} \includegraphics[width=0.32\linewidth]{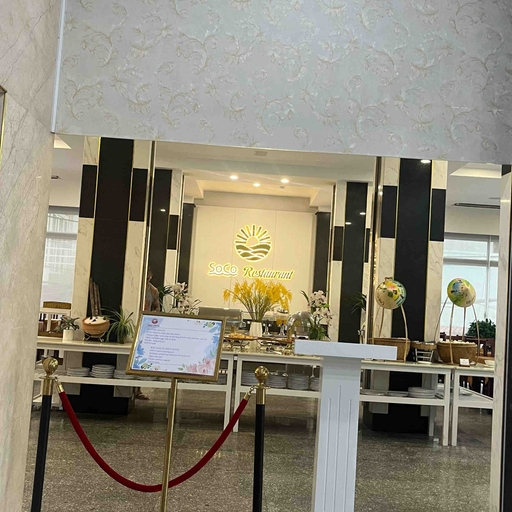}
\includegraphics[width=0.32\linewidth]{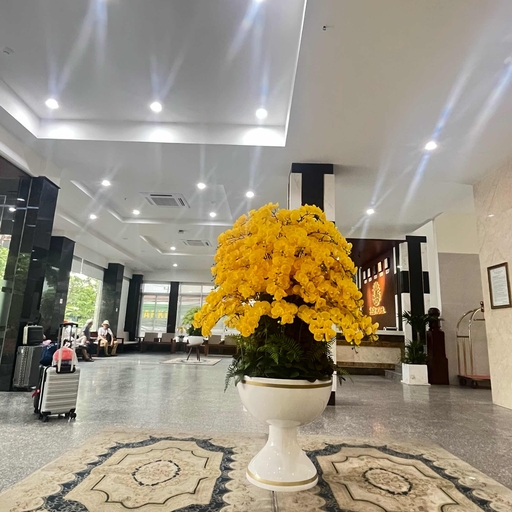}

&  &  &  &  

\\ \hline
% \end{tabular}}
% \end{table}

% \begin{table}[H]
% \centering
% \resizebox{\columnwidth}{!}{
% \begin{tabular}{|p{8cm}|c|c|c|c|}
% \hline
Khách sạn đẹp nhân viên phục vụ chu đáo tận tình Chỉ có ăn sáng giá cao quá mà thức ăn không có gì nhiều. 

&       % ========== Human label ==========

\multirow{3}{*}{\begin{tabular}[c]{@{}l@{}} \\ \\ \\ \\ \\ \\  Food: Negative\\ Room: Positive\\ Service: Positive\\ Public area: Positive\end{tabular}} 

&       % ========== FCMF ==========
\multirow{3}{*}{\begin{tabular}[c]{@{}l@{}} \\ \\ \\ \\ \\ \\  Food: \textcolor{red}{Negative} \cmark \\ Room: \textcolor{red}{Positive} \cmark\\ Service: \textcolor{red}{Positive} \cmark\\ Public area: \textcolor{red}{Positive} \cmark\end{tabular}} 

&       % ========== EF-CapTrRoBERTa ========== 
\multirow{3}{*}{\begin{tabular}[c]{@{}l@{}} \\ \\ \\ \\ \\ \\  Food: Neutral \xmark\\ Room: None \xmark\\ Facilities: Positive \xmark\\ Service: \textcolor{red}{Positive} \cmark \\ Public area: \textcolor{red}{Positive} \cmark\end{tabular}} 

&       % ========== LCF-RoBERTa ==========
\multirow{3}{*}{\begin{tabular}[c]{@{}l@{}} \\ \\ \\ \\ \\ \\  Food: Neutral \xmark\\ Room: None \xmark\\ Facilities: Positive \xmark\\ Service: \textcolor{red}{Positive} \cmark\\ Public area: \textcolor{red}{Positive} \cmark\end{tabular}} \\

 \textit{ \textcolor{blue}{The hotel is beautiful, and the staff are attentive and friendly. However, the breakfast is limited, and the prices are expensive. }}

&  &  &  &  \\
\includegraphics[width=0.32\linewidth]{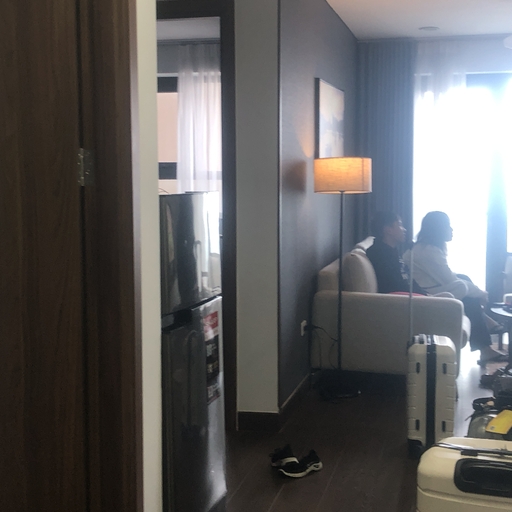} \includegraphics[width=0.32\linewidth]{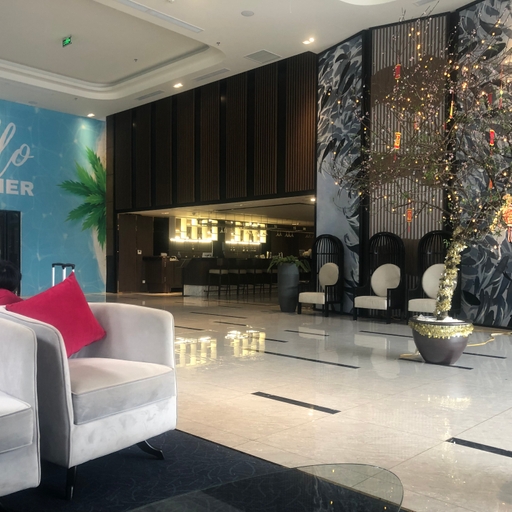}
\includegraphics[width=0.32\linewidth]{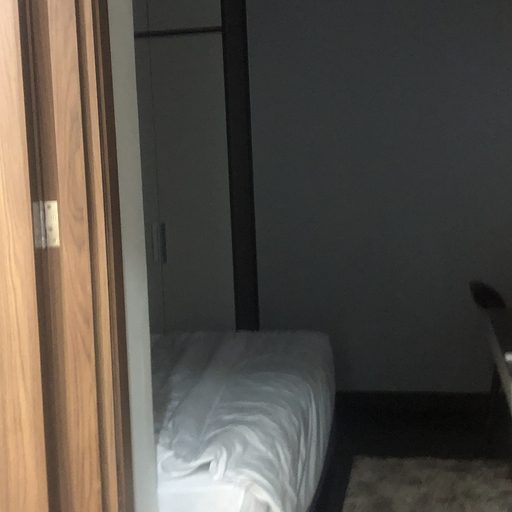} 

\includegraphics[width=0.32\linewidth]{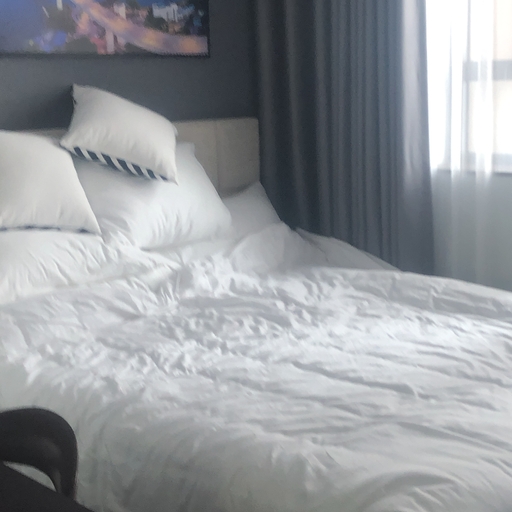} \includegraphics[width=0.32\linewidth]{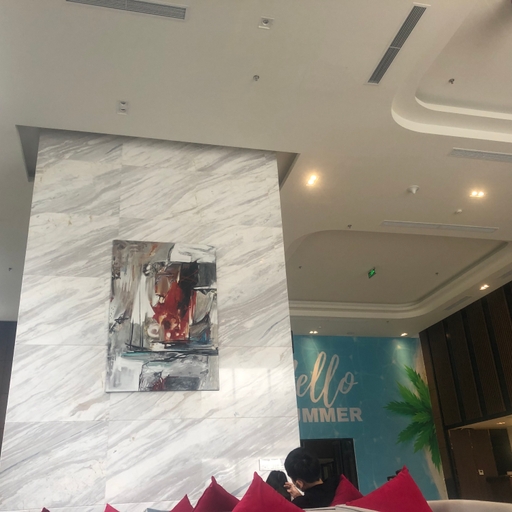}

 &  &  &  &  \\ \hline
\end{tabular}}
\end{table}

% We select several representative test samples to compare the predictions of different methods, 

Table \ref{tab:inc_pred} illustrates examples of mispredictions for each type of error. These errors fall into three main categories:
\begin{itemize}
    \item Failed to identify explicit aspect: The model did not recognize the word "anh bell" (which means "Bellman" in English) as a hotel employee. This is because the user translated the word "anh bell" word-for-word from English ("Mr. Bell"), which is a common practice on social media tools. Consequently, the model failed to predict the "Service" aspect.
    
    \item Mismatched sentiment towards identified implicit aspect: The model correctly predicted "Room" as the implicit aspect in image 1 but failed to recognize the positive sentiment.  The sentence "Tôi đã lưu trú ở đây lần thứ 2 và thấy hài lòng khi ở đây" ("This is my second stay, and the experience has consistently impressed me") clearly indicates a positive experience, which the model overlooked.
    
    \item Misidentified aspects in images lead to misidentified implicit aspects: The model incorrectly predicted the aspect in the image, leading to the failure to identify the implicit aspect mentioned.
\end{itemize}

\begin{table}[H]
\centering
\caption{Error Examples in FCMF framework. \text{\cmark} and \text{\xmark}  denote correct and incorrect predictions.}
\label{tab:inc_pred}
\resizebox{\columnwidth}{!}{%
\begin{tabular}{|l|p{9cm}|l|l|}
\hline
Error Cases & \multicolumn{1}{c|}{Input} & \multicolumn{1}{c|}{Human Label} & \multicolumn{1}{c|}{Prediction} \\ \hline

% =========== CASE 1 ===========
\multirow{5}{*}{\begin{tabular}[c]{@{}l@{}}\\ \\ \\ \\ \\ \\ \\ \\ \\ \\ \\ \\Failed to identify \\  explicit aspect \\   \end{tabular}} \hspace{0.9cm} 
&
1 trải nghiệm quá kinh khủng, vừa mới tới, gặp \textbf{\textcolor{red}{anh bell}} mặt rất quạo, ở cái phòng với cái view là bãi rác, đặt 2 đêm mà không thể ở nổi, nên phải hủy. hình phòng vệ sinh quảng cáo có bồn tắm nằm, nhưng tới là cái tolet nhỏ xíu. trong phòng thì gài khách hàng bằng cách chỉ để 2 chai nước suối trong tủ lạnh, nhà mình lấy uống, thì lại báo tính tiền, nhưng lại không hề để nước suối miễn phí. rẩt là tức, và không bao giờ muôna quay lại &

\multirow{5}{*}{\begin{tabular}[c]{@{}l@{}} \\ \\ \\ \\ \\ \\ \\ \\ \\ \\ Image 1: Public area\\ Room: Negative\\ Service: Negative\\ Public area: Negative\end{tabular}} &
  
\multirow{5}{*}{\begin{tabular}[c]{@{}l@{}}\\ \\ \\ \\ \\ \\ \\ \\ \\ \\ Image 1: Public area \cmark \\ Room: Negative \cmark \\ Service: \textcolor{purple}{None} \xmark \\ Public area: Negative \cmark \end{tabular}} \\
&                            &                                   &                                 \\
&
 
\textit{\textcolor{blue}{Absolutely terrible experience. I just arrived and met \textbf{\textcolor{red}{Bellman}} with a very grumpy face. The room I got had a view of the garbage dump. I booked for two nights but could not stay, so I had to cancel. The advertised bathroom photo showed a bathtub, but it was a tiny toilet when I arrived. The room also tricked guests by only providing two water bottles in the fridge. When we drank them, we were charged, but there was no free water provided. Very angry, and will never come back again.}}

& & \\ &  &  & \\ &   
\includegraphics[width=0.3\linewidth]{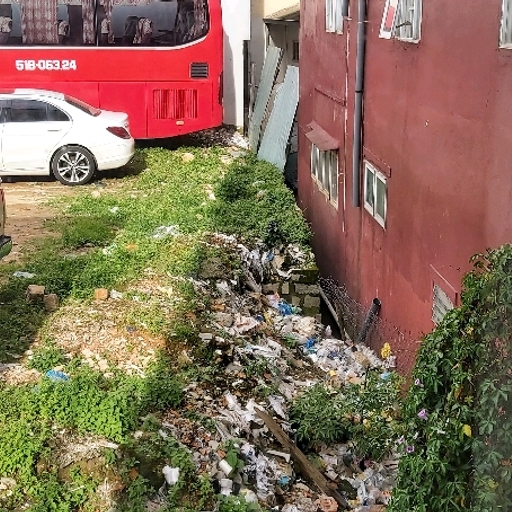} 
&                              &                                 \\ \hline

% \end{tabular}%
% }
% \end{table}

% \begin{table}[h]
% \centering
% \resizebox{\textwidth}{!}{%
% \begin{tabular}{|l|p{11cm}|l|l|}
% \hline
% Error Cases & \multicolumn{1}{c|}{Input} & \multicolumn{1}{c|}{Human Label} & \multicolumn{1}{c|}{Prediction} \\ \hline

% =========== CASE 2 ===========
\multirow{5}{*}{\begin{tabular}[c]{@{}l@{}}\\ \\ \\ \\ \\ \\ \\ Mismatched sentiment \\ towards identified \\ implicit aspect \\   \end{tabular}}
 &
Khách sạn ở trung tâm thành phố, đi lại tham quan rất tiện lợi. Nhân viên khách sạn nhiệt tình, thân thiện. Buffet sáng đa dạng món ăn. \textbf{\textcolor{red}{Tôi đã lưu trú ở đây lần thứ 2 và thấy hài lòng khi ở đây.}} &

\multirow{5}{*}{\begin{tabular}[c]{@{}l@{}} \\ \\ \\ \\ \\ \\ \\  Image 1: Room\\ Location: Positive\\ Food: Positive\\ Room: Positive\\ Service: Positive\end{tabular}} &

\multirow{5}{*}{\begin{tabular}[c]{@{}l@{}} \\ \\ \\ \\ \\ \\ \\  Image 1: Room \cmark\\ Location: Positive \cmark\\ Food: Positive \cmark\\ Room: \textcolor{purple}{Neutral} \xmark\\ Service: Positive \cmark\end{tabular}} \\
&                            &                                   &                                 \\
&

\textit{\textcolor{blue}{The central location of this hotel makes it very convenient to explore all the main attractions of the area. The staff are friendly and helpful, and the breakfast buffet features various dishes. \textbf{\textcolor{red}{This is my second stay, and the experience has consistently impressed me.}}
}}
& & \\&   &   &  \\  &                     
\includegraphics[width=0.3\linewidth]{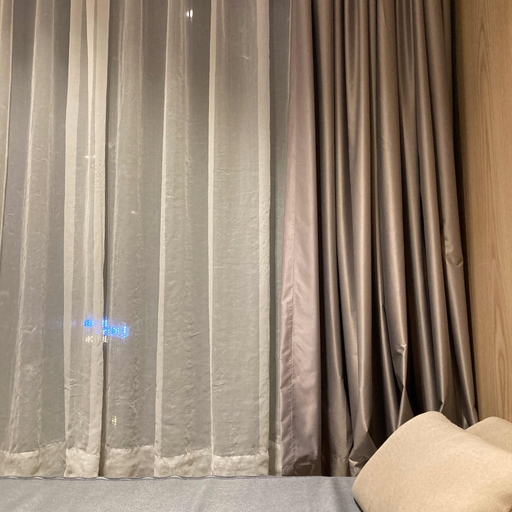} 
&                                   &                                 \\ \hline

% =========== CASE 3 ===========
\multirow{5}{*}{\begin{tabular}[c]{@{}l@{}}\\ \\ \\ \\ \\ \\ \\ \\ \\ \\ Misidentified aspects \\ in images lead  to\\  misidentified implicit \\  aspects   \\ \end{tabular}}
& 
khách sạn ở cách Sun world 3,5km. Đi taxi hết 50k. khách sạn 3 sao nhưng phòng đạt chuẩn 5 sao. gối nằm cực sướng, không bị đau cổ. Bàn chải mềm, chăn ấm nệm êm. Giường 2m rộng rãi. Lễ tân niềm nở dễ thương. ăn sáng buffet hơi ít món nhưng cũng ổn. Gần khách sạn có nhiều quán ăn. Đáng để ở lắm. Mình rất ưng ý. &

\multirow{5}{*}{\begin{tabular}[c]{@{}l@{}} \\ \\ \\ \\ \\ \\ \\ \\ \\ \\ Image 1: Public area\\ Location: Positive\\ Food: Neutral\\ Room: Positive\\ Service: Positive\\ Public area: Positive\end{tabular}} &
  
\multirow{5}{*}{\begin{tabular}[c]{@{}l@{}} \\ \\ \\ \\ \\ \\ \\ \\ \\ \\ Image 1: \textcolor{purple}{Room} \xmark\\ Location: Positive \cmark \\ Food: Neutral \cmark \\ Room: Positive \cmark \\ Service: Positive \cmark \\ Public area: \textcolor{purple}{None} \xmark \end{tabular}} \\
&                            &                                   &                                 \\
 &
 
\textit{\textcolor{blue}{The hotel is located 3.5 km from Sun World. It costs 50k to take a taxi. The hotel is 3-star, but the room is 5-star. The pillows are extremely comfortable, with no neck pain. Soft toothbrush, warm blanket, soft mattress. The 2m bed is spacious. The receptionist is friendly and lovely. The breakfast buffet is a bit small but okay. There are many restaurants near the hotel. Worth staying. I am very satisfied.}}

&
& \\ &  &  &   \\ & 
\includegraphics[width=0.3\linewidth]{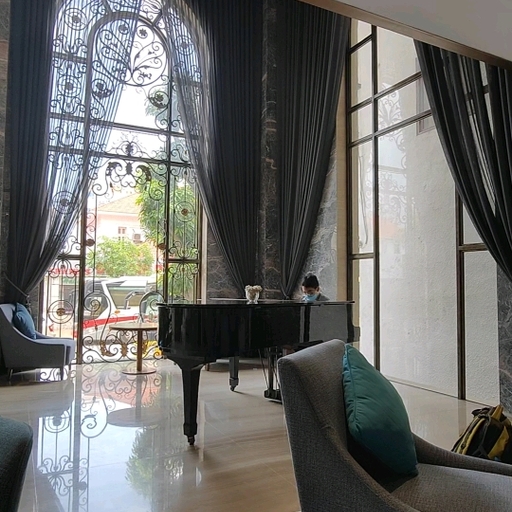} 
&                                   &                                 \\ \hline
\end{tabular}%
}
\end{table}

\subsection{Visualization of Aggregating Image Information Related to Comments} \label{sec:visual_img_rv}

\begin{figure}[h]
    \centering
    \includegraphics[width=1.0\linewidth,height=8cm, keepaspectratio]{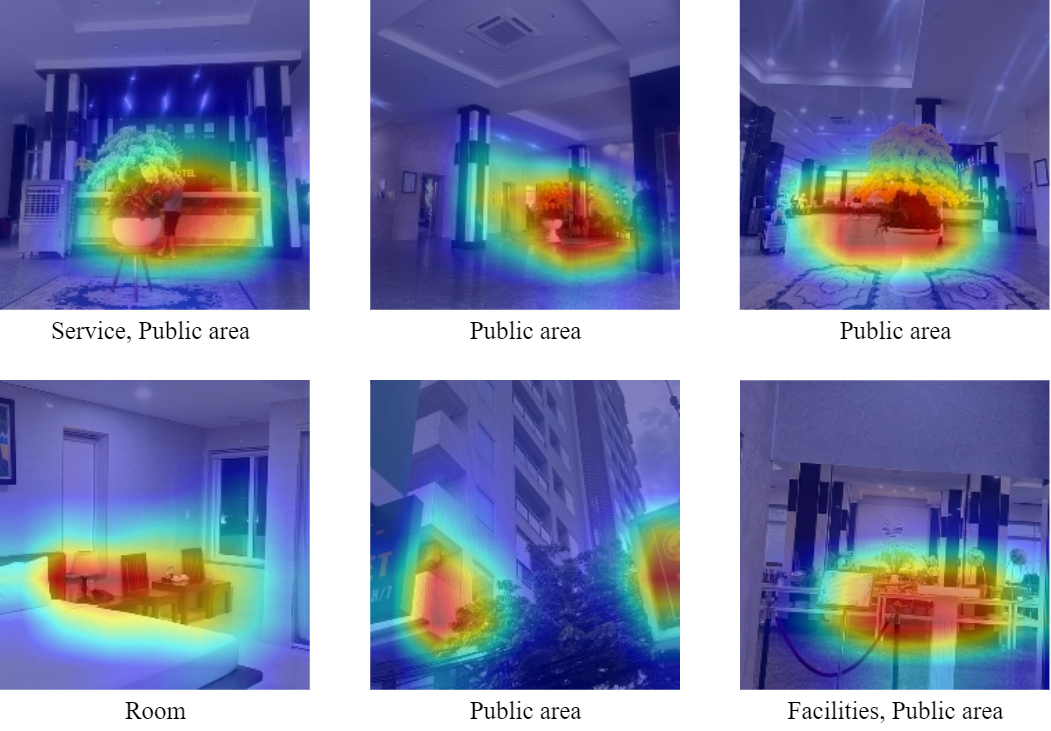}
    \caption{Grad-CAM for Aspect Category Detection task in each image. The aspect category is displayed below each image.}
    \label{fig:visual_attn}
\end{figure}

To understand ResNet's decision-making process and identify important regions, we use Grad-CAM (Gradient-weighted Class Activation Mapping) to visualize the regions within images that significantly influence predictions for the aspect category detection task in the first row of Table \ref{tab:case-study}, as shown in Figure \ref{fig:visual_attn}. In the first image, the ResNet model focuses on the employee and the reception desk to predict the corresponding aspect categories of "Public area" and "Service". In the final image, the ResNet model focuses on the food area of the hotel's buffet restaurant to predict the aspect categories "Facilities" and "Public area". This demonstrates that the ResNet model can accurately focus on relevant image regions to make prediction results.

\begin{figure}[H]
    \centering
    \includegraphics[width=1.0\linewidth]{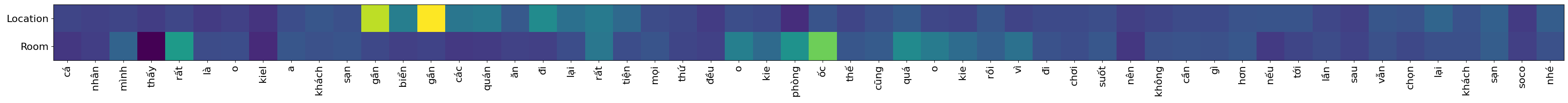}
    \caption{Words related to the Location and Room aspects are color-coded based on their importance in embedding, with brighter colors representing more important words.}
    \label{fig:word_importance}
\end{figure}

The importance of words in the FCMF framework when making explicit aspect predictions "Location" and "Room" is demonstrated in Figure \ref{fig:word_importance}. Our framework pays high attention to the phrase "gần biển gần các quán ăn đi lại rất tiện" (which means "It is conveniently located close to the beach, restaurants" in English) when making predictions for Location, and the phrase "phòng ốc thế cũng quá okie rồi" (which means "The rooms are more than acceptable for my needs" in English) when making predictions for Room. This indicates that our framework can focus on important words and effectively solve explicit aspect prediction tasks.

\begin{figure}[H]
    \centering
    \includegraphics[width=1.0\linewidth,height=10cm, keepaspectratio]{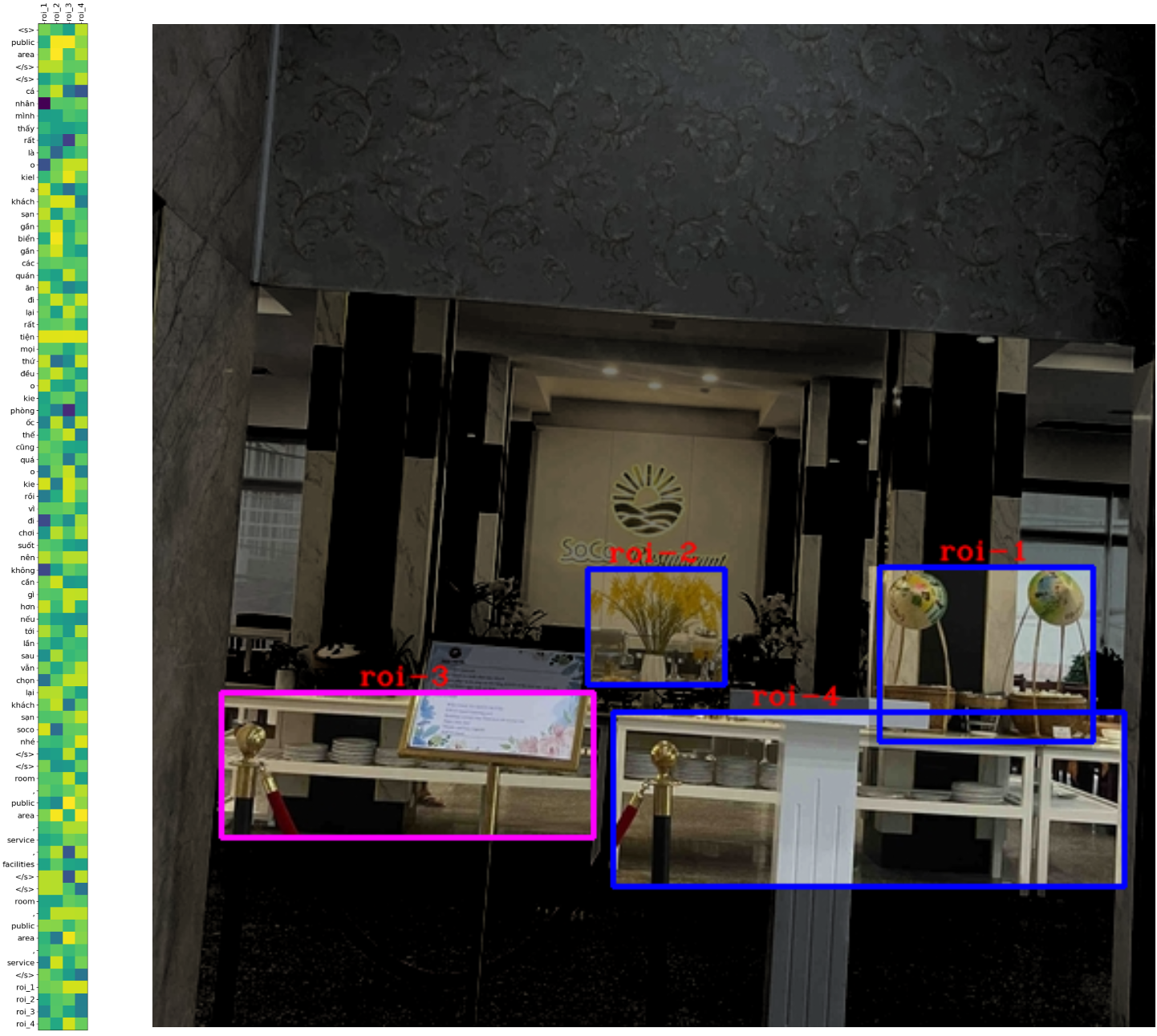}
    \caption{A visualization of Attention scores for each RoI and word in the Geometric RoI-aware module. The brightness of RoIs in the image is proportional to the attention weight of RoI-3.}
    \label{fig:self_attn_roi}
\end{figure}

% note: link tren case study, them aspect duoi anh
% note: bo dau _ 
To better understand the relationship between RoIs, we visualize the attention scores of RoI-3 in relation to other RoIs and the context, as shown in Figure \ref{fig:self_attn_roi}. Our framework determines that RoI-3 is highly correlated with RoI-1 and RoI-4, which represent the food area of a buffet restaurant in a hotel. From this information, our framework can infer that the implicit aspect categories are 'Public area' and 'Facilities.'

\section{Conclusion and Future Works}\label{sec:conclusions}

In this article, we propose ViMACSA - a new Vietnamese multimodal dataset for the Aspect Category Sentiment Analysis task. This dataset consists of 4,876 text-image pairs with fine-grained annotations for both text and image in the hotel domain. Next, we propose a new framework called Fine-Grained Cross-Modal Fusion that effectively learns both intra- and inter-modality interactions between fine-grained textual and visual elements. Experimental results show that our framework achieves remarkable results, outperforming other SOTA models on the proposed dataset. However, we believe that there is still room for improvement. We also delve into the challenges and characteristics of Vietnamese multimodal sentiment analysis, specifically addressing common Vietnamese language errors such as misspellings, abbreviations, and the complexities of Vietnamese language processing. This work contributes both a benchmark dataset and a new framework that leverages fine-grained multimodal information to improve multimodal aspect-category sentiment analysis, supporting future MABSA research.

In future work, we plan to expand our dataset for multi-domain applications to mitigate the effects of domain shift and ensure long-term model robustness. We also plan to incorporate Explanation for Sentiment into our dataset to investigate the relationship between textual elements and specific sentiment expressions, and to study how diverse elements across modalities trigger different sentiments.

\section*{Acknowledgement}
This research was supported by The VNUHCM-University of Information Technology’s Scientific Research Support Fund.

\section*{Data Availability}

Data will be made available on reasonable request.

\section*{CRediT authorship contribution statement}
 \textbf{Quy Hoang Nguyen:} Conceptualization; Data curation; Formal analysis; Investigation; Methodology; Validation; Visualization; Writing - original draft. \textbf{Minh-Van Truong Nguyen:} Conceptualization; Data curation; Formal analysis; Investigation; Validation; Visualization; Writing - review \& editing. \textbf{Kiet Van Nguyen:} Conceptualization; Formal analysis; Investigation; Methodology; Validation; Supervision; Writing - review \& editing. 

\section*{Declaration of Competing Interest}
The authors declare that they have no known competing financial interests or personal relationships that could have appeared to influence the work reported in this article.

% \section*{Acknowledgement}
% This research was supported by The VNUHCM-University of Information Technology's Scientific Research Support Fund.

\appendix
\section{Characteristics of the ViMACSA dataset} \label{sec:char_data}

Our ViMACSA dataset is collected from social media platforms. Therefore, it clearly has the characteristics of online Vietnamese text data. Special characters, misspellings/meaningless words, and abbreviations are all prevalent features of the reviews.

\begin{figure}[H]
    \centering
    \includegraphics[width=1.0\linewidth]{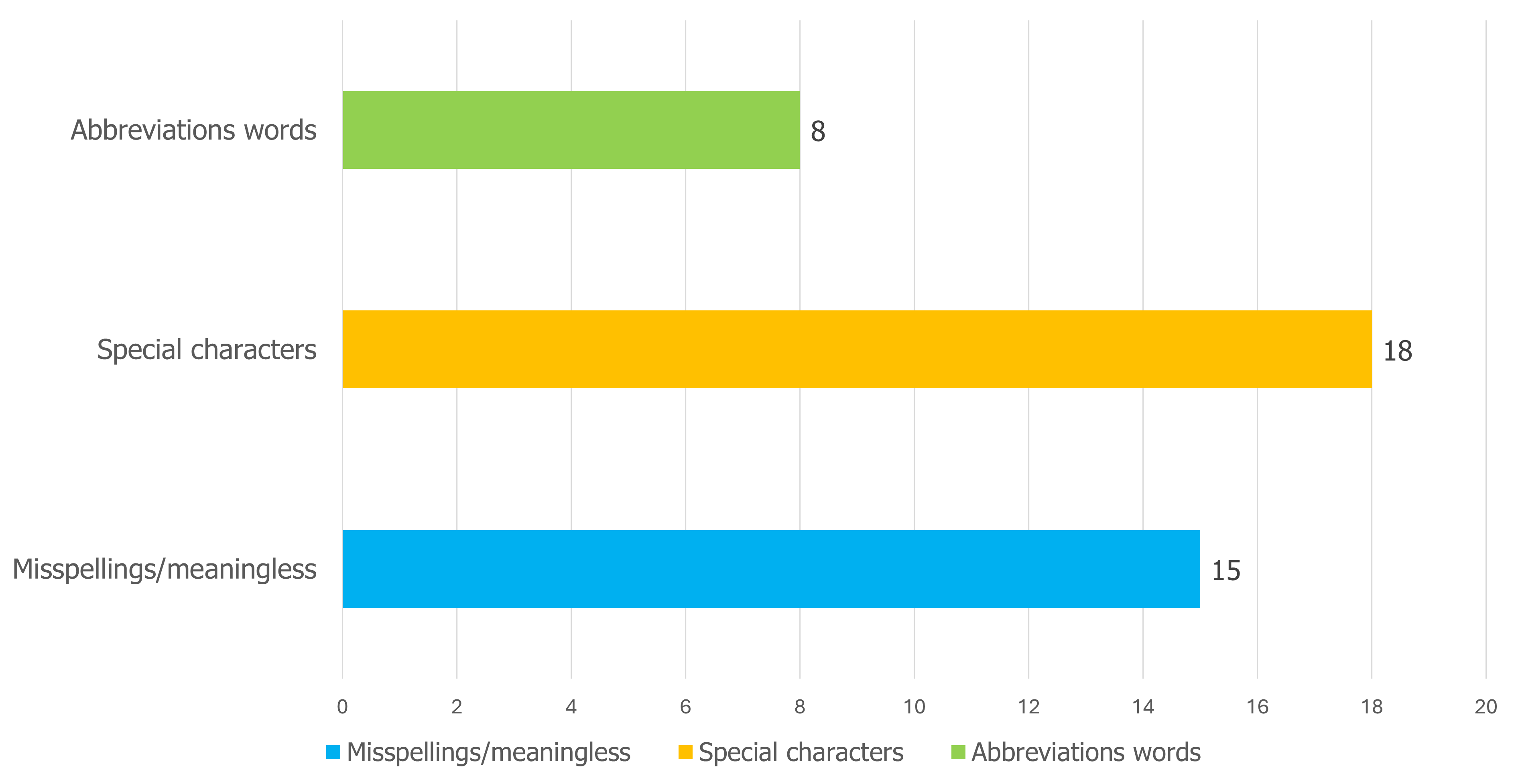}
    \caption{Characteristics statistics on 100 random data points in our dataset.}
    \label{fig:char_dist}
\end{figure}

According to the statistical chart based on 100 random samples from our dataset, as shown in Figure \ref{fig:char_dist}, misspell and meaningless words are the most frequent type of error, followed by special characters and abbreviations. It can be seen that misspellings, abbreviations, and special characters are pretty common on social media. Table \ref{tab:data_error} provides a list of common errors in our dataset, including the following causes:

\begin{itemize}
    \item Incorrect diacritics or words with similar pronunciation: hổ trợ (hỗ trợ), siu (siêu), etc.
    \item Incorrect word lengthening: đẹppppp (đẹp), ngonnnm (ngon), etc.
    \item Abbreviations can be ambiguous due to multiple meanings: ks/ksan (khách sạn), ko/kh (không), etc.
\end{itemize}

\begin{table}[H]
\centering
\caption{Features often appear in the ViMACSA dataset.}
\label{tab:data_error}

\begin{tabular}{|l|l|l|}
\hline
Spelling errors & Abbreviations & Special characters \\ \hline

% spell
\begin{tabular}[c]{@{}l@{}}
hổ trợ $\to$ hỗ trợ    
\\ đẹppppp $\to$ đẹp   
\\ siu $\to$ siêu    
\\ ngonnn $\to$ ngon
\end{tabular} 

% abbre
& \begin{tabular}[c]{@{}l@{}}
Ks $\to$ khách sạn    
\\ ksan $\to$ khách sạn   
\\ gđ $\to$ gia đình    
\\ mn $\to$ mọi người 
\\ đc $\to$ được
\\ ko $\to$ không 
\\ kh $\to$ không 
\end{tabular} 

% emoji
& \begin{tabular}[c]{@{}l@{}}
:)    
\\ :D    
\\ =))
\\ Emoji: \emojione, \emojitwo, \emojithree
\end{tabular}

\\ \hline
\end{tabular}
\end{table}

% preprocess

Therefore, these data are diverse and have characteristics that do not follow conventional rules. Before conducting the experiments, we performed text preprocessing with the following specific steps: 

\begin{itemize}
    \item Lowercase Sentences: Lowercasing ensures consistency across the data, making it easier for machine learning algorithms to process and analyze the text.

    \item Delete Redundant Space and Characters: Redundant spaces and characters can clutter data and make it harder to read and analyze. Removing redundancies promotes uniformity and simplifies analysis. 

    \item Convert Unicode: To ensure consistent text representation across the data, we converted text to UTF-8.

    \item Word Segmentation: In most languages, words are separated by spaces, but in Vietnamese, it presents a unique challenge because there are no clear visual boundaries between words. This characteristic is further compounded by the morphological complexity of Vietnamese, where words are often formed by adding prefixes and suffixes to root morphemes. Text segmentation serves as a crucial step in addressing this challenge, as it enables the identification of these meaningful linguistic units for subsequent analysis.

\end{itemize}

\begin{figure}[H]
    \centering
    \includegraphics[height=5cm]{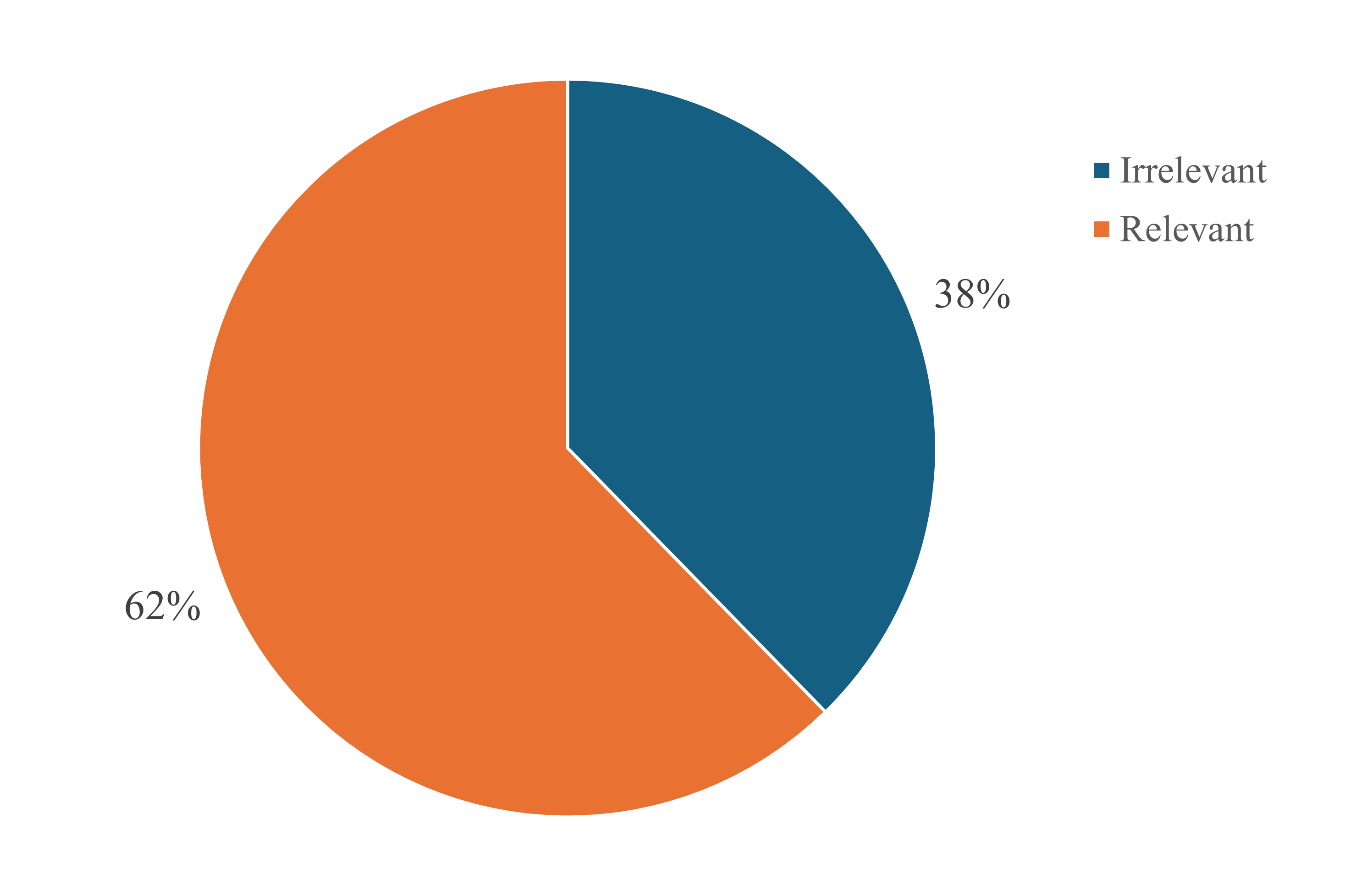}
    \caption{Ratio of relevant/irrelevant images in the ViMACSA dataset.}    
    \label{fig:pie_image}
\end{figure}

Figure \ref{fig:pie_image} illustrates the proportion of relevant and irrelevant images in the ViMACSA dataset. The proportion of irrelevant images is relatively high, up to 38\%. This suggests that users on social media platforms tend to share many images that are not directly related to their experience.

% TODO: chỉnh lại tên hình
\section{Vocabulary} \label{sec:vocal_wl}

\begin{figure}[H]
    \centering
    \includegraphics[width=1.0\linewidth]{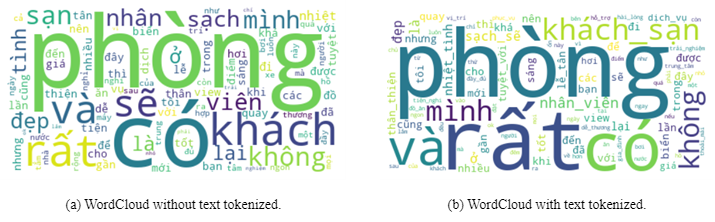}
    \caption{Word cloud representation of ViMACSA dataset.}
    \label{fig:tokenize}
\end{figure}

Figure \ref{fig:tokenize} represents two word clouds for the reviews within our dataset. Each cloud comprises one hundred words, with Cloud B explicitly generated from the tokenized text. In this section, we focus mainly on the tokenized word cloud. It is worth noting that word segmentation is a crucial step in Vietnamese language processing as it lacks inherent word boundaries. This process ensures accurate word identification and helps gain a deeper comprehension of the text.

The most common words in the cloud are related to aspects of the hotel mentioned in the reviews. Several notable nouns are "khách sạn" (hotel), "phòng" (room), "lễ tân" (receptionist), "nhân viên" (staff). In addition, the adverb "rất" is used frequently, which is similar to the English word "very". There are also some adjectives such as "thân thiện" (friendly), mainly used to describe the staff, and "tuyệt vời" (wonderful), often used to express positive emotions.

\section{Detail information of Annotation Guidelines}
\label{sec:detail_annotation} 
% and sentiment polarity 
This section includes detailed information on the definition of aspect category in the ViMACSA dataset. 
\begin{itemize}
    \item Room: This category includes comments on the hotel room, covering aspects such as size, design, furniture, bathroom, soundproofing, and more.
    \item Location: This category refers to the hotel's location, nearby attractions, and more. 
    \item Food: This category includes feedback on breakfast, food, drinks, the buffet, and more.
    \item Facilities: This category includes the amenities provided by the hotel to meet the individual needs and interests of guests, such as the swimming pool, gym, restaurant, internet, spa, club, children's amenities, and more.
    \item Service: This category covers general comments on the service, staff attitude, quality of service, any issues related to check-in and check-out, and more. 
    \item Public Area: This category refers to the view and common spaces available to guests, including the lobby, hallway, garden, courtyard, and more.
\end{itemize}

% \begin{figure}[H]
%     \centering
%     \includegraphics[width=1.0\linewidth,height=7cm,keepaspectratio]{horizon_cross_attn_image.png}
%     \caption{Attention scores for each word correspond to different image regions in the image-guided module. Higher brightness indicates greater attention scores.}
%     \label{fig:cross_attn_image}
% \end{figure}

\bibliographystyle{model1-num-names}
%\bibliographystyle{cas-model2-names}

\bibliography{cas-refs}


% \bio{MrKietNguyen}
% Kiet Van Nguyen received the B.S. and M.S. degrees from the VNUHCMC - University of Information Technology in 2012 and 2017, respectively. He is currently a lecturer at the Faculty of Information Science and Engineering, VNUHCMC - University of Information Technology in Ho Chi Minh City, Vietnam. His research interests include natural language processing, machine reading comprehension, and deep learning.

% \bio{DrNganNguyen}
% Ngan Luu-Thuy Nguyen received the Ph.D. degree in information science and technology from the University of Tokyo, Japan. She was a Postdoctoral Researcher with the National Institute of Informatics, Japan, from 2012 to 2013. She is currently a Scientist at the VNUHCMC - University of Information Technology in Ho Chi Minh City, Vietnam. Her research interests include natural language processing and data analysis.

\end{document}